\documentclass[dvipsnames,format=sigconf,anonymous=false, review=false]{acmart}

\AtBeginDocument{%
  }

\setcopyright{acmlicensed}

\copyrightyear{2025}
\acmYear{2025}
\setcopyright{acmlicensed}
\setcctype{by}
\acmConference[GECCO '25]{Genetic and Evolutionary Computation
Conference}{July 14--18, 2025}{Malaga, Spain}
\acmBooktitle{Genetic and Evolutionary Computation Conference (GECCO '25),
July 14--18, 2025, Malaga, Spain}\acmDOI{10.1145/3712256.3726391}
\acmISBN{979-8-4007-1465-8/2025/07}

\usepackage{caption}
\usepackage{algorithm}
\usepackage{algpseudocode}
\usepackage{subcaption}
\usepackage{graphicx}
\usepackage{seqsplit}
\usepackage{array}
\usepackage{makecell}
\usepackage{color}
\usepackage{soul}
\usepackage{comment}
\usepackage{amsmath}
\usepackage{tabularx}
\usepackage{booktabs}
\usepackage{threeparttable}
\usepackage{siunitx}  
\usepackage{float}
\usepackage{colortbl}
\usepackage{xcolor}
\usepackage{url}

\algnewcommand\algorithmicforeach{\textbf{for each}}
\algdef{S}[FOR]{ForEach}[1]{\algorithmicforeach\ #1\ \algorithmicdo}

\begin{document}

\title{A Perturbation and Speciation-Based Algorithm\\ for Dynamic Optimization Uninformed of Change}

\author{Federico Signorelli}
\orcid{https://orcid.org/0009-0003-4210-0798}
\affiliation{%
  \institution{Computer Science Department}
  \city{Amsterdam}
  \country{The Netherlands}
}
\email{f.r.signorelli@student.vu.nl}

\author{Anil Yaman}
\orcid{https://orcid.org/0000-0003-1379-3778}
\affiliation{%
  \institution{Computer Science Department}
  \city{Amsterdam}
  \country{The Netherlands}
}
\email{a.yaman@vu.nl}

\renewcommand{\shortauthors}{Signorelli and Yaman}

\begin{abstract}
Dynamic optimization problems (DOPs) are challenging due to their changing conditions. This requires algorithms to be highly adaptable and efficient in terms of finding rapidly new optimal solutions under changing conditions. Traditional approaches often depend on explicit change detection, which can be impractical or inefficient when the change detection is unreliable or unfeasible. We propose Perturbation and Speciation-Based Particle Swarm Optimization (PSPSO), a robust algorithm for uninformed dynamic optimization without requiring the information of environmental changes.
The PSPSO combines speciation-based niching, deactivation, and a newly proposed random perturbation mechanism to handle DOPs. PSPSO leverages a cyclical multi-population framework, strategic resource allocation, and targeted noisy updates, to adapt to dynamic environments. 
We compare PSPSO with several state-of-the-art algorithms on the Generalized Moving Peaks Benchmark (GMPB), which covers a variety of scenarios, including simple and multi-modal dynamic optimization, frequent and intense changes, and high-dimensional spaces.  Our results show that PSPSO outperforms other state-of-the-art uninformed algorithms in all scenarios and leads to competitive results compared to informed algorithms. In particular, PSPSO shows strength in functions with high dimensionality or high frequency of change in the GMPB. The ablation study showed the importance of the random perturbation component.
%
\end{abstract}



\maketitle

\section{Introduction}

Traditional optimization techniques often assume a static environment where the objective function remains constant over time. However, many real-world problems are inherently dynamic with changing conditions that necessitate continuous adaptation of the solution \cite{mavrovouniotis_survey_2017}. These problems, known as dynamic optimization problems (DOPs)\cite{branke_memory_1999}, aim to maximize or minimize a function changing in time. We focus on DOPs with discrete changes over time without constraints~\cite{branke_opt} that can be formalized as: 
\[ \text{Maximize:} \]
\[
  f^{(t)}(x) = \{ f(x, \alpha^{(k)}) \}_{k=1}^T = \{ f(x, \alpha^{(1)}), f(x, \alpha^{(2)}), .., f(x, \alpha^{(T)}) \}
\]
\[\text{and} \ x = \{x_1, x_2, \ldots, x_D \}\]
\[
\text{Subject to:} \ x \in X : X = \{x \ | \ Lb_i \leq x_i \leq Ub_i \}, \ i \in \{1, 2, \ldots, D \}
\]
where \( f \) is the objective function, t $\in \{1, \ldots, T\}$ is the time index, \( x \) is a solution within the \( D \)-dimensional bounded search space \( X \), and \( \alpha(t) \) is a vector of time-dependent control parameters \cite{yazdani_species-based_2023,yazdani_survey_2021}.

Metaheuristic methods inspired by nature, such as Evolutionary Algorithms (EAs) \cite{bartz2014evolutionary} and Swarm Intelligence (SI) \cite{chakraborty2017swarm} algorithms, are derivative-free, stochastic, population-based, global search algorithms \cite{weise2009global} used for a wide range of optimization problems, including static, large-scale, constrained, and multi-objective optimization \cite{michalewicz1996evolutionary, ho2004intelligent, coello2007evolutionary, fonseca1995overview}. Their population-driven and iterative characteristics are particularly shown to be well suited for DOPs \cite{mavrovouniotis_survey_2017, nguyen_evolutionary_2012}.  
%
However, metaheuristic methods that work with static problems need to address the changing characteristics of DOPs .

One naive approach for dealing with the DOPs could be reinitialization after a change in the fitness function~\cite{branke_opt, hatzakis2006dynamic}. This approach may not be the most effective since it disregards possibly important information on the search process (i.e., perhaps previously found solutions could be used), and furthermore, relies on the information of change.

Other approaches achieved increasingly better success with DOPs by proposing Dynamic Optimization Algorithms (DOA) that make use of techniques in metaheuristics. These techniques include~\cite{yazdani_survey_2021}: convergence detection \cite{du2013differential, yazdani2013novel, trojanowski2009properties, nguyen2013solving}, explicit archiving \cite{halder_self-adaptive_2011, branke_memory_1999, wang2007triggered}, diversity control \cite{branke2000multi, blackwell_multi-swarm_2004, parrott2004particle, trojanowski1999searching, blackwell2007particle, blackwell2006multiswarms}, population division and management \cite{blackwell_multi-swarm_2004, blackwell2006multiswarms, li_adaptive_2016}, change detection \cite{branke_memory_1999, blackwell2006multiswarms, hu_2002_detection}.

Convergence detection methods, relying on fitness difference \cite{du2013differential, yazdani_2013_novel} or spatial threshold \cite{blackwell2006multiswarms} tracking, are often adopted in combination with diversity control methods such as removal \cite{li_clustering_2009}, deactivation \cite{kamosi_2010_hibernating, novoa_2010_improvement, yazdani_2013_novel} or randomization \cite{blackwell2007particle, danial_yazdani_scaling_2020} of converged populations, with the aim of saving computational resources to use them more effectively. The use of an explicit archive has mostly been discarded in favour of this combination of methods.
Population division and management methods rely on the strategy applied also to multi-modal optimization, which addresses problem spaces with several peaks (modes) \cite{das2011real}, of using multiple independent subpopulations to track multiple optima. This idea holds up well for DOPs because monitoring multiple promising areas can enhance the chance of finding a new global optimum following an environmental shift, making the use of these methods wide spread in state-of-the-art DOAs \cite{yazdani_survey_2021}.

Most of the proposed approaches rely on the information of the changes in the problem \cite{boulesnane2021we, yazdani_survey_2021}. While in certain real-world DOPs this information may be available~\cite{nguyen2011continuous}, in most scenarios, this is not a realistic assumption~\cite{li_general_2012}. Many other algorithms make use of a change detection mechanism~\cite{boulesnane2021we} based on fitness monitoring or reevaluation-based methods. Fitness monitoring-based methods aim to detect a change in the environment through a change in fitness values of all \cite{nasiri_2016_firefly} or best particles~\cite{hu_2002_detection}. This approach may not always be reliable~\cite{nguyen_2012_challenges, richter_2009_detecting}. Reevaluation-based methods make use of detector particles that are reevaluated frequently to detect changes \cite{branke_memory_1999, hu_2002_detection}; they appear to be more reliable and to guarantee precision if a sufficient number of detectors are used \cite{richter_2009_detecting, yazdani_survey_2021}. Environmental changes appear to be particularly elusive to track when they are subtle or partial within the search space \cite{li_general_2012}, as well as when noise is present in the environment \cite{li2008benchmark}. Thus, it is crucial to devise algorithms that do not rely on the information of the change or these detection mechanisms.

In this work, we address this challenge by introducing an algorithm that does not rely on the information of changes. The aim is to both improve the performance of dynamic optimization in environments uninformed of the change, and to identify simple and key mechanisms that can produce state-of-the-art results compared to uninformed algorithms as well as competitive results compared to informed algorithms~\cite{yazdani2020benchmarking}. We introduce a new algorithm referred to as Perturbation and Speciation-based Particle Swarm Optimization (PSPSO), which takes inspiration from several predecessors CPSOR \cite{li_general_2012}, AMSO \cite{li_adaptive_2014} and AMPPSO \cite{li_adaptive_2016} in terms of its use multi-population and cyclic deactivation and removal of subpopulations through overlap and convergence detection. These ideas were combined with the proposed speciation mechanism \cite{parrott2004particle} to create guided subpopulations, which demonstrated its success in IDSPSO \cite{blackwell_particle_2008} and SPSO\_AP\_AD \cite{yazdani_species-based_2023}. In addition to this novel combination of established components, our algorithm introduces a noisy perturbation component that can impact both activated and deactivated subpopulations, with the goal of enhancing robustness at every iteration. The combination of all components introduces an effective synergy to tackle DOPs without the information of change.

We test our algorithm on the Generalized Moving Peaks Benchmark (GMPB), which provides state-of-the-art benchmarking problems in dynamic optimization, and compare our algorithm with state-of-the-art algorithms proposed for DOPs. These algorithms include CPSOR \cite{li_general_2012}, AMPPSO \cite{li_adaptive_2016}, DynDE ~\cite{dynDE}, DSPSO \cite{parrott_locating_2006}, AMSO \cite{li_adaptive_2014} from the uninformed category, and mCMAES \cite{danial_yazdani_scaling_2020}, ACFPSO \cite{yazdani_adaptive_2022}, SPSO\_AP\_AD \cite{yazdani_species-based_2023}, AmQSO \cite{blackwell_particle_2008}, IDSPSO \cite{blackwell_particle_2008} from the informed category. Our results show the effectiveness of our algorithm. In comparison with the uninformed ones, our results provide state-of-the-art results, and in comparison with the informed algorithms, provide competitive performance across all scenarios. In particular, the PSPSO shows strengths in scenarios with frequent change, high-dimensional, and with large shifts. 
We further perform an ablation study of the proposed component and sensitivity analysis for the parameter assignments to show that the proposed component plays a crucial role in its success. 

\section{Related Work}

This section reviews commonly used mechanisms in DOAs that aim to track moving optimum~\cite{nguyen_2012_challenges, yazdani_survey_2021, yazdani_species-based_2023}, clustering for multi-population generation~\cite{blackwell2006multiswarms, blackwell2007particle, li_general_2012}, particle removal and deactivation~\cite{li_clustering_2009, li_adaptive_2014, yazdani_adaptive_2022}, and adaptive components \cite{li_adaptive_2014, yazdani_species-based_2023}.



\noindent{}\textbf{Tracking the moving optimum.} Most DOAs emphasize tracking the moving optimum (TMO)~\cite{nguyen_2012_challenges, yazdani_survey_2021, yazdani_species-based_2023, danial_yazdani_scaling_2020, li_adaptive_2016, lung_2007_tracking, yang_2010_tracking}. This approach focuses on localizing and following peaks in dynamic optimization instead of just maintaining diversity to prevent convergence. Self-Organizing Scouts (SOS) \cite{branke2000multi} is a pioneering method that achieves TMO using multiple populations. SOS employs two categories of subpopulations: one for exploration and discovery of local optima, and several smaller subpopulations for exploiting and tracking each promising region. This framework has been integrated into numerous multi-population algorithms with various adaptations.

\noindent{}\textbf{Unsupervised clustering approaches.} The DynDE algorithm~\cite{dynDE} involves multiple fixed subpopulations to tackle dynamic optimization. After initialization of subpopulations, each is assigned a Differential Evolution (DE)~\cite{lampinen2005differential} variant by an allocating scheme which can also be stochastic. If the best individuals of two subpopulations are close, the subpopulation with the inferior solution is reset. 
%


An adaptive multi-population DE algorithm with clustering is proposed in \cite{halder_self-adaptive_2011}. The algorithm starts by clustering individuals using the K-means algorithm. Each cluster improves its own members through the "DE/best/1" scheme. To maintain effective search and prevent overlap, clusters are periodically re-evaluated based on performance. Satisfactory performance is determined if the number of changes in the global best value exceeds a predefined threshold. If performance is satisfactory, the cluster number is reduced; if it is poor, a new cluster is introduced to maintain diversity. It employs environmental change detection using a test particle (i.e., solution) that continuously monitors changes in its objective value, prompting a restart of the algorithm when a change is detected. To utilize previous knowledge effectively, the algorithm employs an external archive where the best particles of converged clusters are stored and reintroduced after environmental changes are detected.

The Crowding Archive in cluster-based Crowding Differential Evolution (CDE) \cite{mukherjee_2014_crowding} works similarly, but utilizes fuzzy C-means clustering to form and refine clusters of individuals. The algorithm also begins by clustering individuals and improving them using intra-cluster mutation, and clusters are re-evaluated, and re-clustered based on performance to adapt to changes. Environmental changes are detected with a test solution, triggering re-initialization. The external archive is here defined as a Crowding Archive, which retains the best solutions from clusters and uses them after changes to help track optima.

Cellular PSO \cite{hashemi_cellular_2009} uses cellular automata to manage particle swarms in dynamic environments. Each particle is assigned to a cell, and local best positions are updated based on neighboring cells. Particles update their velocities and positions to explore the search space. If a cell's particle density exceeds a threshold, some particles are reinitialized to maintain diversity. The algorithm resets memory upon detecting environmental changes, ensuring continuous adaptation and efficient exploration.


Clustering Particle Swarm Optimization (CPSO) \cite{li_clustering_2009} addresses the challenges of guiding particles to promising areas and determining the optimal number of subpopulations. CPSO employs a global search strategy combined with single-linkage hierarchical clustering to form effective sub-swarms. Initially, a global search is performed to distribute the population across the search space. Rough clustering is then applied, grouping particles based on the shortest distance between any two members, in an attempt to make the subpopulation division dynamic and suited to the search space. subpopulations have a maximum size, and the clusters are refined throughout the run by merging overlapping clusters based on clusters' radii and a defined distance ratio. The algorithm updates the global best by integrating promising information from improved particle dimensions, while every subpopulation except the best gets deactivated once it converges, to save computational resources.
A test particle continuously monitors for changes in the objective value, triggering a restart of the algorithm upon detection. During this process, the convergence list, which records the best particles from converged sub-swarms (like the Crowding Archive), seeds the new cradle swarm that is reintroduced in the population together with random individuals in response to a change.

CPSOR \cite{li_general_2012} builds on the CPSO by eliminating the need for change information. It similarly initializes using single-linkage hierarchical clustering and applies mechanisms for detecting overlap and preventing overcrowding based on a maximum subpopulation size. As in CPSO, local search enhancement involves checking each dimension to improve the global best. While CPSOR retains CPSO's diversity mechanism, it introduces diversity not by detecting change but by monitoring active individuals. When the percentage of deactivated or removed individuals out of the total initial ones goes below a diversity threshold, the algorithm reintroduces the best solutions from the crowding archive and random individuals into the population.

\noindent{}\textbf{Speciation.} 
SPSO \cite{li_adaptive_2016} introduced the idea of clustering using speciation, in which the clusters are created around species seeds based on ranking. SPSO is improved in \cite{li_particle_2006}  by employing convergence detection and subpopulation removal.  
DSPSO \cite{parrott_locating_2006} does not use the information of the changes. The algorithm introduces additional features to enhance performance in changing fitness landscapes. 
A key modification in DSPSO is its capability to track moving optima. Each iteration, the algorithm re-evaluates the fitness of each particles' individual best position. This allows particles to use current fitness information while retaining spatial knowledge from earlier environmental states. 
Additionally, to encourage exploration and prevent convergence at known optima, DSPSO implements a maximum species population parameter. If the population within a species exceeds this limit, individuals with lower fitness are reinitialized randomly. IDSPSO \cite{blackwell_particle_2008} improves the efficiency of reinitialization in speciation after an environmental change, which requires, however, a detection mechanism. 

\noindent{}\textbf{Adaptive components.} Several works focused on implementing adaptive components in terms of parameter tuning or multi-population control~\cite{blackwell_particle_2008,li_adaptive_2014,huang_survey_2020, brest_2009_adaptive, ibrahim_2021_wind, yazdani_adaptive_2022}. Adaptive mQSO (AmQSO) \cite{blackwell_particle_2008} is an adaptive multi-population method that dynamically adjusts the number of subpopulations based on the number of discovered peaks. AmQSO begins with a single subpopulation. Once this subpopulation converges to a peak, a new subpopulation is created and initialized. Given the unknown number of peaks, the algorithm continually searches for undiscovered peaks by initializing new subpopulations. Unlike its predecessor mQSO\cite{blackwell2006multiswarms}, which uses quantum particles throughout the optimization process to maintain local diversity within each subpopulation, AmQSO employs quantum particles only immediately after environmental changes to diversify each subpopulation and address local diversity loss. Changes are detected through use of test particles and this approach prevents the unnecessary expenditure of computational resources on maintaining local diversity over time. Furthermore, AmQSO determines the exclusion radius by using the number of subpopulations as an estimate of the number of peaks. While this estimation is highly sensitive to the accuracy of the convergence detection method and may be prone to errors, the idea of an adaptive exclusion mechanism has been widely adopted in the design of DOP algorithms.

AMSO \cite{li_adaptive_2014} follows up on the ideas of CPSO and CPSOR with another uninformed multi-population algorithm. AMSO still relies on single-linkage hierarchical clustering to create the different subpopulations, crowd control with merging subpopulations at overlap and then trimming if their size exceeds a threshold, convergence detection and removal of the subpopulation, and a diversity mechanism that when triggered introduces random individuals and best individuals saved from each converged subpopulation in a temporary archive. The key difference with predecessor CPSOR is that in the diversity mechanism the number of populations and the moments to increase diversity are adaptive. The moment to trigger diversity is based on the drop rate of subpopulations, with the idea that when subpopulations stop overlapping or converging, that means the algorithm is stagnating and diversity should be introduced. The number of individuals introduced with diversity is increased or decreased in the same direction as the trend of active individuals of population in time.

The Adaptive Multi-Population (AMP) framework \cite{li_adaptive_2016} aims to provide 
a versatile, effective approach for dynamic optimization which does not rely on the information of the changes and has main mechanisms of clustering, tracking and adapting. Clustering uses single linkage hierarchical clustering and requires no set parameter since the stop condition relies on a comparison between inter and intra-cluster distance. This makes it so that the subpopulation size is variable, but also that a minimum or maximum size cannot be guaranteed. 
AMP also tries to detect stagnant subpopulations, which have not converged and therefore maintain a large radius. While local search is enhanced for the best particle through Brownian movement, Cauchy movement is employed on stagnant subpopulations to help them converge. 
A diversity mechanism is triggered when the average radius of non-stagnating populations falls below a certain threshold, and consists in awakening the deactivated converged subpopulations as well as an introduction of new individuals which will undergo the clustering procedure. Importantly, the number of total individuals is variable and is updated adaptively throughout the run using a probabilistic prediction scheme informed by historical data, with the number of individuals involved in the search growing proportionally with the number of peaks found. AMPPSO leverages this framework with PSO as its backbone optimizer.

In the Adaptive Control Framework (ACF) \cite{yazdani_adaptive_2022}, subpopulations are categorized into explorers, exploiters, and trackers based on their radius, which reflects their convergence status. Initially, all subpopulations start as explorers, randomly distributed across the search space to identify peaks. When the diversity of an explorer falls below a threshold \(R\), it becomes an exploiter, focused on moving towards the peak summit. Once an exploiter reaches close proximity to the peak, indicated by its diversity falling below a smaller threshold \(r\), it transitions to a tracker, tasked with tracking the peak’s movement and providing information on peak shift severity and number of optima. ACF uses a double-layer exclusion mechanism to manage subpopulations, with two thresholds \(E\) and \(e\) to prevent overlap and redundant resource usage. subpopulations that enter the exclusion zone of another are re-initialized or removed based on their type and performance. Additionally, ACF employs an adaptive resource allocation strategy, prioritizing active subpopulations based on their role and performance. Upon detecting environmental changes, ACF increases the diversity of trackers according to the estimated shift severity and re-evaluates all stored solutions to update memory. ACFPSO utilizes the adaptive control framework with PSO as its optimizer.

SPSO\_AP\_AD \cite{yazdani_species-based_2023} improves speciation niching methods by introducing adaptive population and deactivation mechanisms. After being initially divided into subpopulations through the process of speciation, using ranked list and species heads and adding a fixed number of closest individuals to each subpopulation, species proceed with their search and are classified as trackers (converged to a local optimum) or non-trackers (exploring), optimizing resource allocation. 
The adaptive deactivation component saves computational resources by deactivating converged tracker species, with the deactivation radius adjusted based on the amount of tracker populations. To maintain diversity, new random individuals are injected when all species converge, and an exclusion mechanism prevents overcrowding by removing redundant species.
The change reaction component here does not only attempt to detect the changes but also measures shift severity by calculating Euclidean distances between tracker species' best positions across environments. Non-seed members are reinitialized around seed positions based on estimated shift severity to address local diversity loss. Stored solutions are reevaluated to update the swarm's memory.
To prevent over-exploitation, a maximum species threshold $N_{max}$ is enforced, reinitializing individuals of the worst species when exceeded. The workflow involves starting with a randomized population and continuously adapting species formation, deactivation, and population size based on environmental changes, enhancing local diversity, and updating stored solutions as needed.

\section{Methods}
This section introduces our proposed algorithm\footnote{The code of this paper is publicly accessible here: \url{https://github.com/FreddyDeWatersir/PSPSO\#}}, PSPSO, and discusses the mechanisms it uses. 





\noindent{}\textbf{General outline of the algorithm.} The outline of the algorithm is provided in Algorithm~\ref{alg:flow}. First, a population of solutions is randomly initialized and divided into $n$ subpopulations (subswarms) with a size of $s$ solutions through the process of speciation. 
The process of speciation ensures that each subswarm remains relatively cohesive and focuses on exploring distinct areas of the search space.

\begin{algorithm}
\caption{Outline of the PSPSO algorithm }
\label{alg:flow}
\begin{algorithmic}[1]
\State Initialization
\State Speciation based niching \Comment{to form subpopulations}

\While {stopping criteria is not met} \Comment{Main loop}
    \ForEach {active subpopulation $i$}
        \State subswarm $i$ update
    \EndFor
    \ForEach {pair $(i,j)$ of subpopulations }
        \State overlap detection
    \EndFor
    \State subswarm perturbation
    \ForEach {active subpopulation $i$}
        \State convergence detection
    \EndFor
    \If { $n_{active} / n s < \alpha$} 
        \State diversity mechanism
    \EndIf
\EndWhile

\end{algorithmic}
\end{algorithm}

\noindent{}\textbf{Swarm update.} Each subswarm is updated based on the Particle Swarm Optimization (PSO). In this setup, the global best position (\( \mathbf{g}_{\text{best}} \)) for a subswarm corresponds to the best position found by that subswarm over time. Notably, the best positions from other subswarms are not shared or accessible, ensuring that the search remains localized to the subswarm's region. This localized search strategy is particularly effective for multi-modal optimization problems, where identifying multiple optima in a complex fitness landscape is the goal.

The behavior of each particle within a subswarm is governed by the velocity update rule \cite{kennedy_1995_PSO}:
\begin{equation}
\label{eq:swarmUpdate}
\mathbf{v}_i^{(t+1)} = w \mathbf{v}_i^{(t)} 
+ c_1 r_1 \big( \mathbf{p}_{\text{best},i} - \mathbf{x}_i^{(t)} \big) 
+ c_2 r_2 \big( \mathbf{g}_{\text{best}} - \mathbf{x}_i^{(t)} \big),
\end{equation}
where \( w \) is the inertia weight controlling the influence of the particle's previous velocity, \( c_1 \) and \( c_2 \) are cognitive and social acceleration coefficients, and \( r_1 \) and \( r_2 \) are random values in $[0,1]$ sampled from a uniform distribution. The term \( \mathbf{p}_{\text{best},i} \) represents the particle's personal best position, while \( \mathbf{g}_{\text{best}} \) is the global best position within the same subswarm.

Once the velocity is updated, the particle's position is adjusted accordingly as:
\begin{equation}
\label{eq:update2}
\mathbf{x}_i^{(t+1)} = \mathbf{x}_i^{(t)} + \mathbf{v}_i^{(t+1)}.
\end{equation}
Particles are then ensured to be within search bounds $[Lb, Ub]$.

\noindent{}\textbf{Overlap detection.} An inefficient use of resources may occur when two distinct subswarms explore the same area of the search space. To address this issue, we make use of a parameter-free method to detect overlap between two subpopulations. Then, when overlap is detected between two subpopulations, the worst one is removed. 
 
To detect overlapping subpopulations, Euclidean distance $\text{dist}_{i,j}$ between the (subswarm) global best positions of each pair of subpopulations $i$ and $j$ is computed.

Two subpopulations are considered overlapping if:
\[
\text{dist}_{i,j} < r_i \quad \text{and} \quad \text{dist}_{i,j} < r_j,
\]
\[
\text{where } r_i = \frac{1}{s} \sum_{k=1}^{s} \|\mathbf{x}_k - \mathbf{c}_i\| \quad \text{and} \quad \mathbf{c}_i = \frac{1}{s} \sum_{k=1}^{s} \mathbf{x}_k.
\] $r_i$ is the initial radius of subpopulation $i$ defined in the speciation process as the mean distance of its particles from its center $c_i$.

If two subpopulations overlap, the one with the inferior global best fitness value is removed.


\noindent{}\textbf{Subswarm perturbation.} In every iteration, a subpopulation is selected randomly with uniform probability to undergo a perturbation step as follows:
\[
\mathbf{v}_j^{(new)} = \mathbf{v}_j^{(old)} + \mathbf{r}, \quad \mathbf{r} \sim \mathcal{U}(-P, P)^D, \quad 
P = p  (Ub - Lb)
\]
where $\mathbf{v}_j$ is the velocity of particle $\mathbf{x}_j \in \mathbb{R}^D$ and $j$ is the position of particle $P$. $P$ is the perturbation range, $p$ is the perturbation factor and $U_b$, $L_b$ are the upper and lower bounds of the search domain. The perturbation step provides also the possibility for deactivated subpopulations to be updated.


\noindent{}\textbf{Convergence detection.} The goal of this mechanism is to detect convergence of subpopulations and initiate deactivation to save resources, unless the subpopulation is the global best (i.e., containing the best individual globally). To detect convergence as follows:
\[
\text{if } R_i < R, \quad \text{where} \quad R_i = \frac{1}{s} \sum_{j=1}^{s} \sqrt{\sum_{k=1}^D \left(p_{j,i,k} - c_{i,k}\right)^2}
\]
where $R = 0.01D$, and $D$ indicates the dimensionality of the problem.

If a subswarm is converged, it is deactivated. This means that the individuals of the subpopulation are not removed but excluded from regular subswarm fitness and velocity updates during every iteration. Deactivation can help saving resources while keeping the individuals for possible further improvements.


\noindent{}\textbf{Diversity mechanism.} 
%
%
At every iteration, we calculate the ratio between the number of active individuals $n_{active}$ in the population and the initial total amount of individuals $n s$ (e.g., this can change based on the deactivation and removal steps). If this ratio drops below the predefined parameter $\alpha\in [0, 1]$, the diversity mechanism is triggered. This leads to the removal of all deactivated subpopulations while storing the best particle from each subpopulation in a temporary archive, which is then used to reintroduce them into the population. To match the initial size of the population, randomly generated particles are introduced into the population. After these steps, the speciation process is triggered to identify subpopulations.  


\noindent{}\textbf{Speciation based niching.} This process takes a population of particles (i.e., their positions, velocities, and fitness scores), and divides into $n$ subpopulations. Initially, all individuals are ranked in order of fitness. The fittest individual in the list is designated as the first species head, which means that it is removed from the ranked list and added to a newly created subpopulation. Subsequently, the Euclidean distances $d(\mathbf{x}_h, \mathbf{x}_j)$ between this head and all individuals left in the ranked list are computed. 
%
%
The $s - 1$ closest individuals are then also removed from the ranked list and added to the head's subpopulation. This is repeated until the ranked list is over; all the subpopulations are formed by going to the updated ranked list, making the fittest individual left head of the new species, then adding the $s - 1$ closest members to its subpopulation. 

This process is utilized not only at the first initialization of the population, but every time the diversity mechanism is activated. Speciation was chosen as the multi-population approach because of its effectiveness in performance and solid theoretical foundation based on the idea that the use of species' heads guides the spread of subpopulations by giving them directionality and enhancing immediate search capacity, which appears to be highly desirable for TMO. We set the parameter $s$ instead of a distance threshold as the way to create subpopulations. In this way we can avoid the known difficulty of having to choose an arbitrary distance threshold parameter in an unknown environment.

\section{Experimental Setup}

To assess the performance of our algorithm we use twelve different benchmark scenarios from the Generalized Moving Peaks Benchmark (GMPB) \cite{benchmark_2024} that are used in the most recent dynamic optimization competitions~\cite{yazdani2024competitiondynamicoptimizationproblems}. Table~\ref{tab:problem-instances} summarizes the properties of these functions.


\begin{table}[H]
    \centering
    \caption{\small Parameter settings of the 12 problem instances in GMPB.}
    \resizebox{0.48\textwidth}{!}{
    \begin{tabular}{|c|c|c|c|c|}
        \hline
        \textbf{GMPB Scenario} & \textbf{Number of Peaks} & \textbf{ChangeFrequency} & \textbf{Dimensions} & \textbf{ShiftSeverity} \\ \hline
        F1 & 5  & 5000 & 5  & 1 \\ \hline
        F2 & 10 & 5000 & 5  & 1 \\ \hline
        F3 & 25 & 5000 & 5  & 1 \\ \hline
        F4 & 50 & 5000 & 5  & 1 \\ \hline
        F5 & 100 & 5000 & 5  & 1 \\ \hline
        F6 & 10 & 2500 & 5  & 1 \\ \hline
        F7 & 10  & 1000 & 5  & 1 \\ \hline
        F8 & 10  & 500  & 5  & 1 \\ \hline
        F9 & 10  & 5000 & 10 & 1 \\ \hline
        F10 & 10 & 5000 & 20 & 1 \\ \hline
        F11 & 10 & 5000 & 5  & 2 \\ \hline
        F12 & 10 & 5000 & 5  & 5 \\ \hline
    \end{tabular}}
    \label{tab:problem-instances}
\end{table}

The performance of the algorithms is measured using offline error \cite{branke_designing_2003} (\(E_{O}\)), which is established in the literature as a common and reliable metric \cite{yazdani_survey_2021}. \(E_{O}\) aims to evaluate the ability of the algorithm to readily search the dynamic environment by calculating the average error of the best found position over all fitness evaluations using the following equation:

\[
E_O = \frac{1}{T\vartheta} \sum_{t=1}^{T} \sum_{c=1}^{\vartheta} \left( f^{(t)}(x^{\star (t)}) - f^{(t)} \left( x^{\ast ((t-1)\vartheta + c)} \right) \right)
\]

where \( x^{\star (t)} \) is the global optimum position at the \( t \)-th environment, \( T \) is the number of environments, \( \vartheta \) is the change frequency, \( c \) is the fitness evaluation counter for each environment, and \( x^{\ast ((t-1)\vartheta + c)} \) is the best found position at the \( c \)-th fitness evaluation in the \( t \)-th environment \cite{yazdani_survey_2021}.

\begin{table}[]
    \centering
    \begin{threeparttable}
    \caption{PSPSO Parameter Settings}
    \label{tab:parameterSetup}
    \begin{tabular}{|l|l||c|}
        \hline
        \textbf{Name} & \textbf{Symbol}  & \textbf{Value} \\ \hline
        Inertia & $w$ & 0.6 \\ \hline
           Personal and social Coefficient & $c_1, c_2$ & 2.83 \\ \hline
        Swarm Size & s & 7 \\ \hline
        Number of Swarms       & $n$ & 10 \\ \hline
        Diversity Rate Threshold & $\alpha$ & 0.7 \\ \hline
        Radius for Convergence & $R$ & $0.01  D$ \tnote{*} \\ \hline  
        Perturbation Range       & $P$ & $ 0.025 (Ub -Lb) $ \tnote{**}\\ \hline
    \end{tabular}
    \begin{tablenotes}
        \footnotesize
        \item[*] $D$ is the number of dimensions.
        \item[**] $Ub$ and $Lb$ are upper and lower bound of search space.
        
    \end{tablenotes}
    \end{threeparttable}
    
\end{table}

\noindent{}\textbf{Compared algorithms.} We choose 10 state-of-the-art algorithms from two categories, informed and uninformed in terms of changes, to compare our proposed approach. The algorithms in the informed category include: mCMAES \cite{danial_yazdani_scaling_2020}, ACFPSO \cite{yazdani_adaptive_2022}, SPSO\_AP\_AD\cite{yazdani_species-based_2023}, AmQSO \cite{blackwell_particle_2008}, IDSPSO \cite{blackwell_particle_2008}; and the algorithms in the uninformed category include: DynDE~\cite{dynDE}, CPSOR \cite{li_general_2012}, DSPSO \cite{parrott_locating_2006}, AMSO \cite{li_adaptive_2014}, AMPPSO \cite{li_adaptive_2016}. Note that our algorithm is in the uninformed category and that all the informed algorithms are here explicitly informed when an environmental change happens. 

\noindent{}\textbf{Parameter settings.} Table~\ref{tab:parameterSetup} summarizes the parameters used in our algorithm. The parameters $w, c_1$ and $c_2$ of the PSO optimizer were taken from the most recent implementation of CPSOR \cite{peng_edolab_2024, li_general_2012}. All parameters specific to PSPSO were established through sensitivity analysis, already informed by past sensitivity analysis conducted for CPSOR \cite{li_general_2012}. An ablation study was also undertaken around the novel perturbation mechanism to verify its utility.

\begin{table*}[]
    \centering
    \captionsetup{justification=centering}
    \caption{Offline errors and standard errors of different algorithms on the GMPB. The results indicated with bold face are the best within the uninformed algorithms, and the results highlighted gray are the best overall in both uninformed and informed algorithms ($\alpha=0.05$). No statistical significance between results that are highlighted within the same column.}
    \label{tab:allResults}
    \resizebox{1\textwidth}{!}{
    \begin{tabular}{|l|c|c|c|c|c|c|c|c|c|c|c|c|}
        \hline
        \textbf{Uninformed Algorithms} & \textbf{F1} & \textbf{F2} & \textbf{F3} & \textbf{F4} & \textbf{F5} & \textbf{F6} & \textbf{F7} & \textbf{F8} & \textbf{F9} & \textbf{F10} & \textbf{F11} & \textbf{F12}\\ \hline
        PSPSO & \cellcolor{lightgray}\textbf{1.63(0.17)}& \textbf{2.31(0.10)} & \textbf{4.13(0.14)}& \textbf{4.26(0.15)} & \textbf{4.43(0.15)} & \textbf{2.90(0.15)} & \cellcolor{lightgray}\textbf{3.51(0.13)}& \cellcolor{lightgray}\textbf{5.41(0.16)} & \cellcolor{lightgray}\textbf{5.64(0.33)} & \cellcolor{lightgray}\textbf{20.82(2.03)} & \textbf{2.79(0.13)} & \textbf{4.64(0.13)}\\ \hline
        CPSOR & 22.17(1.02) & 22.35(0.89) & 18.60(0.70) & 17.57(0.46) & 16.28(0.38) & 18.58(0.61) & 17.88(0.70) & 19.95(0.99) & 63.66(3.26) & 177.90(4.87) & 26.07(0.73) & 34.82(0.72) \\ \hline
        AMSO & 13.82(0.55)  & 12.83(0.74) & 12.21(0.57) & 12.54(0.43) & 12.16(0.41) & 11.44(0.33) & 13.84(0.56) & 16.45(0.53) & 46.43(2.85) & 148.62(7.44) & 19.07(0.79) & 28.07(0.81)\\ \hline
        DSPSO & 31.10(2.01) & 30.29(1.56) & 25.15(0.90) & 21.32(0.65) & 21.40(0.74) & 28.91(1.28) &29.23(1.54) & 30.11(1.45) & 117.40(0.90) & 248.12(1.57) & 32.36(1.16) & 40.12(0.71) \\ \hline
        AMPPSO & 3.13(0.12) & 3.61(0.10) & \textbf{4.33(0.12)} & \textbf{4.41(0.08)} & \textbf{4.40(0.08)} & 4.95(0.17) & 7.72(0.26) & 10.63(0.29) & 10.09(0.31) & 35.37(1.34) & 4.74(0.10) & 7.69(0.13) \\ \hline
        DynDE  &  38.34(2.07)&  37.54(1.18)&  31.73(1.13)&  28.15(0.92)& 27.25(0.97)& 35.93(1.67)& 39.99(1.77)& 39.15(1.80)& 111.87(3.03)& 244.91(3.28)& 38.49(1.42)& 47.10(1.02)\\ \hline
        \textbf{Informed Algorithms} &  &  &  &  & & & & & & & &\\ \hline
        mcMAES  & \cellcolor{lightgray}1.65(0.24) & \cellcolor{lightgray}1.64(0.09) & 1.95(0.05) & 2.41(0.07) & 2.49(0.05) & \cellcolor{lightgray}2.20(0.10) & 4.26(0.25) & 7.08(0.33) & 6.68(0.20) & 29.45(1.89) & \cellcolor{lightgray} 2.45(0.11) & 7.19(0.18)\\ \hline
        ACFPSO & \cellcolor{lightgray}1.39(0.09)& \cellcolor{lightgray}1.71(0.08) & 1.98(0.07) & \cellcolor{lightgray}2.13(0.06) & 2.39(0.06) & 2.39(0.09) & 4.39(0.16) & 6.82(0.24) & 10.22(0.74) & 49.59(3.17) & \cellcolor{lightgray} 2.45(0.09) & \cellcolor{lightgray}4.31(0.11)\\ \hline
        SPSO\_AP\_AD & \cellcolor{lightgray}1.44(0.12) & \cellcolor{lightgray}1.50(0.06) & \cellcolor{lightgray}1.67(0.03) & \cellcolor{lightgray}1.99(0.04) & \cellcolor{lightgray}2.08(0.03) & \cellcolor{lightgray}2.03(0.07) & \cellcolor{lightgray}3.28(0.11) & \cellcolor{lightgray}5.38(0.20) & 5.90(0.15) & 28.80(0.75) & \cellcolor{lightgray}2.23(0.06) & \cellcolor{lightgray}4.13(0.09) \\ \hline
        AmQSO & 1.87(0.08) & 1.99(0.05) & 2.04(0.03) & 2.35(0.03) & 2.56(0.03) & 2.63(0.06) & 4.57(0.12) & 7.36(0.18) & 8.14(0.20) & 62.86(2.44) & 2.82(0.06) & 5.07(0.08)\\ \hline
        IDSPSO & 1.87(0.10) & 2.10(0.10) & 3.28(0.11) & 3.50(0.10) & 3.54(0.08) & 3.39(0.19) & 6.49(0.27) & 13.43(0.56) & 92.43(1.41) & 221.88(1.22) & 3.82(0.12) & 9.98(0.16)\\ \hline
    \end{tabular}}
\end{table*}

\section{Experimental Results}

Table~\ref{tab:allResults} shows the results of PSPSO and the comparison algorithms on the GMPB benchmark functions. The results report the average offline errors of 31 independent runs and their corresponding standard errors. The results in bold indicate the best result in the uninformed category for that scenario, while gray highlight of the cell indicates the algorithm that achieved the best overall in both informed and uninformed categories. We use the Mann-Whitney U test \cite{mann_test_1947} with $\alpha = 0.05$ significance level on the results of the 31 runs between the algorithm with the best numerical offline error and the closest performers to verify if there is a statistically significant difference.

Functions F1-F5 are multi-modal environments with relatively simple change parameters and an amount of modes increasing from 5 to 100. PSPSO and AMPPSO outperform all other uninformed algorithms, with PSPSO showing a particular advantage in simpler multi-modal scenarios. Informed algorithms generally have better performance than uninformed ones, with SPSO\_AP\_AD being the best among them. Interestingly, while PSPSO matches informed algorithms in the simpler problem, the increasing number of peaks also increases the performance gap.

F6-F8 are environments with increasing change frequency. PSPSO exhibits a very strong performance. It outperforms all other uninformed methods and is competitive against all informed ones, increasingly so in fast changing problems. While SPSO\_AP\_AD remains the best performing informed algorithm, PSPSO has similar performance to that of other informed methods. In the most extreme scenarios in terms F7 and F8, PSPSO has optimal performance together with SPSO\_AP\_AD. This suggests that PSPSO suits well scenarios with high change frequency, underlining a capacity to quickly adapt.

In the high-dimensional scenarios with regular change parameters F9 and F10, PSPSO showcases its best performance by obtaining the lowest error of all uninformed and informed algorithms. This underlines the effectiveness of PSPSO in high-dimensional problems. In scenarios with higher shift magnitude (F11 and F12), PSPSO matches the performance of the informed algorithms, particularly SPSO\_AP\_AD.

In summary, PSPSO provides best results relative to the compared state-of-the-art algorithms in uninformed category, and competitive results relative to the informed ones. This indicates the effectiveness and efficiency of the mechanisms and their synergy in dealing with the DOPs.

\noindent{}\textbf{Sensitivity analysis.} We conduct a sensitivity analysis on the parameters, namely the number of swarms $n$, swarm size $s$, and perturbation factor $p$, to ascertain their importance and aid in selecting suitable values for our algorithm (see Table~\ref{tab:sensitivityAnalysis}). An ablation study is also conducted on the perturbation mechanism to verify how performance compares without its use (i.e., when $p=0$). Out of the 12 GMPB scenarios previously analysed, 5 that represented each of the different problem types (simple, multi-modal, frequently changing, high-dimensional, high shift magnitude) were selected for a compact sensitivity analysis. To avoid an exploding number of experiments to run, when the focus is on a parameter, the others will remain fixed according to the values in table \ref{tab:parameterSetup}. The best results in each scenario were represented in bold.

\begin{table}[]
    \centering
    \captionsetup{justification=centering}
    \caption{Offline error and (standard errors) for different values of perturbation factor $p$, swarm size $s$ and subpopulation number $n$}
    \label{tab:sensitivityAnalysis}
    \resizebox{1\linewidth}{!}{
    \begin{tabular}{|l|c|c|c|c|c|c|}
        \hline
        $p$ & 0 & 0.01 & 0.025 & 0.05 \\ \hline
        \hline
        F1 & 14.09(0.81) & \textbf{1.50(0.20)} & 1.63(0.17) & 2.10(0.13) \\ \hline
        F3 & 14.96(0.52) & 4.20(0.17) & \textbf{4.13(0.14)} & 4.62(0.14) \\ \hline
        F8 & 14.21(0.71) & \textbf{5.33(0.24)} & 5.41(0.16) & 7.32(0.15)\\ \hline
        F10 & 128.95(6.81)& 23.33(1.68) & 20.82(2.03) & \textbf{20.76(0.57)} \\ \hline
        F12 & 31.89(0.77) & 5.56(0.20) & 4.64(0.13) & \textbf{4.22(0.09)}\\ \hline
        \hline
        pop. size ($s \times n$)  & 5x10 & 7x10 & 10x10 & 7x15 \\ \hline
        \hline
        F1 & 1.87(0.16) & 1.63(0.17) & 1.60(0.13) & \textbf{1.59(0.09)} \\ \hline
        F3 & 4.20(0.13) & 4.13(0.14) & 4.17(0.23) & \textbf{2.83(0.09)}\\ \hline
        F8 & 6.10(0.18) & \textbf{5.41(0.16)} & 6.15(0.12) & 6.28(0.12) \\ \hline
        F10 & 21.89(1.20) & 20.82(2.03) & \textbf{16.22(0.43)} & 19.58(0.92)\\ \hline
        F12 & 4.38(0.15) & 4.64(0.13) & \textbf{3.98(0.08)} & 4.32(0.09) \\ \hline
    \end{tabular}}
\end{table}

In terms of perturbation parameter $p$, tested assignment values led to similar results on the tested functions. More notably, when the perturbation mechanism is not used, the performance of the algorithm suffers drastically.

In terms of population size, there is some variation in terms of the best choice in different problems. In particular, population size 7x10 appears to be the most effective in high-frequency problems, while 10x10 showed most effectiveness in high-dimensional and high shift magnitude problems, and 7x15 is the optimal choice for a multi-modal scenario. All tested parameters performed well on the high-dimensional problem. Overall, it indicates that several choices for these parameter values are viable, but 7x10 and 10x10 lead to the best results on the test problems.

\section{Conclusions}
This paper introduced PSPSO, a novel dynamic optimization algorithm that utilizes multi-populations with speciation-based niching, adaptive resource management through convergence and overlap detection mechanisms, and a novel method of random subpopulation perturbation, without relying on information about environmental changes, which can especially be beneficial for cases where change detection is unreliable or infeasible. The components of the algorithm were combined carefully to lead to a synergy and provide robust and effective results in dynamic problems. We compared our algorithm with the other state-of-the-art algorithms that are informed and uninformed of the environmental change in DOPs.

Our findings showed that PSPSO consistently outperforms other uninformed methods and provides competitive results compared to informed algorithms across all scenarios tested in the GMPB benchmark commonly used in dynamic optimization. In particular, our algorithm shows optimal performance in frequently changing and high-dimensional scenarios. The novel component of random perturbation of a subpopulation is highlighted as the key contributor to the success of the algorithm based on the ablation study. Here, the noisy update of particle velocities demonstrates to be a beneficial heuristic for exploration in dynamic problems.

The algorithm's ability to perform effectively without explicit change detection underscores the potential for developing robust optimization techniques that do not depend on many parameters and a way to obtain information about environmental shifts \cite{li_general_2012}. The trend in improving DOPs effectiveness involves designing algorithms that are highly adaptive and dynamically adjust parameters based on observations collected during execution \cite{yazdani_2021_A}. However, it's crucial to note that these methods, which depend on change detection mechanisms, may face challenges when information is not easily accessible. The proposed approaches in this paper shows the viability and competitiveness of approaches that do not rely on the environment change information.

PSPSO's robust performance in rapidly changing, high-dimensional dynamic environments, likely due to the random perturbation mechanism, suggests intriguing research directions in efficient exploration mechanisms in high-frequency and dimensional dynamic environments. Incorporating noise in the algorithm and experimenting with ways of leveraging deactivated subpopulations also appear as possibly promising avenues of further research, and it appears interesting to verify whether this sort of components can also be combined with state-of-the-art adaptive parameter tuning approaches.

Another noteworthy research direction involves utilizing PSPSO for dynamic real-world problems, such as hyperparameter optimization~\cite{kalita_2020_svm}, dynamic economic dispatch \cite{wang_2011_dispatch}, odor source localization \cite{jatmiko_2007}, and contamination source detection in water \cite{liu2006adaptive, liu2011contamination}. Applying PSPSO to real-world dynamic optimization problems would validate its practical utility and can identify areas for further improvement. 

In conclusion, the development and evaluation of PSPSO highlight the value of designing algorithms that rely on minimal assumptions about environmental changes while maintaining competitive performance across diverse and challenging scenarios. This work enhances the comprehension of optimization algorithms' success in uncertain conditions by focusing on robustness through principled and simple mechanisms. These insights invite further exploration of how randomness, simplicity, and efficient resource management can be systematically combined to improve performance.

\bibliographystyle{ACM-Reference-Format}
\bibliography{literature}


\begin{thebibliography}{68}


\ifx \showCODEN    \undefined \def \showCODEN     #1{\unskip}     \fi
\ifx \showISBNx    \undefined \def \showISBNx     #1{\unskip}     \fi
\ifx \showISBNxiii \undefined \def \showISBNxiii  #1{\unskip}     \fi
\ifx \showISSN     \undefined \def \showISSN      #1{\unskip}     \fi
\ifx \showLCCN     \undefined \def \showLCCN      #1{\unskip}     \fi
\ifx \shownote     \undefined \def \shownote      #1{#1}          \fi
\ifx \showarticletitle \undefined \def \showarticletitle #1{#1}   \fi
\ifx \showURL      \undefined \def \showURL       {\relax}        \fi
\providecommand\bibfield[2]{#2}
\providecommand\bibinfo[2]{#2}
\providecommand\natexlab[1]{#1}
\providecommand\showeprint[2][]{arXiv:#2}

\bibitem[Bartz-Beielstein et~al\mbox{.}(2014)]%
        {bartz2014evolutionary}
\bibfield{author}{\bibinfo{person}{Thomas Bartz-Beielstein},
  \bibinfo{person}{J{\"u}rgen Branke}, \bibinfo{person}{J{\"o}rn Mehnen}, {and}
  \bibinfo{person}{Olaf Mersmann}.} \bibinfo{year}{2014}\natexlab{}.
\newblock \showarticletitle{Evolutionary algorithms}.
\newblock \bibinfo{journal}{\emph{Wiley Interdisciplinary Reviews: Data Mining
  and Knowledge Discovery}} \bibinfo{volume}{4}, \bibinfo{number}{3}
  (\bibinfo{year}{2014}), \bibinfo{pages}{178--195}.
\newblock


\bibitem[Blackwell(2007)]%
        {blackwell2007particle}
\bibfield{author}{\bibinfo{person}{Tim Blackwell}.}
  \bibinfo{year}{2007}\natexlab{}.
\newblock \showarticletitle{Particle swarm optimization in dynamic
  environments}.
\newblock \bibinfo{journal}{\emph{Evolutionary computation in dynamic and
  uncertain environments}} (\bibinfo{year}{2007}), \bibinfo{pages}{29--49}.
\newblock


\bibitem[Blackwell and Branke(2004)]%
        {blackwell_multi-swarm_2004}
\bibfield{author}{\bibinfo{person}{Tim Blackwell} {and}
  \bibinfo{person}{J{\"u}rgen Branke}.} \bibinfo{year}{2004}\natexlab{}.
\newblock \showarticletitle{Multi-swarm optimization in dynamic environments}.
  In \bibinfo{booktitle}{\emph{Workshops on applications of evolutionary
  computation}}. Springer, \bibinfo{pages}{489--500}.
\newblock


\bibitem[Blackwell and Branke(2006)]%
        {blackwell2006multiswarms}
\bibfield{author}{\bibinfo{person}{Tim Blackwell} {and}
  \bibinfo{person}{J{\"u}rgen Branke}.} \bibinfo{year}{2006}\natexlab{}.
\newblock \showarticletitle{Multiswarms, exclusion, and anti-convergence in
  dynamic environments}.
\newblock \bibinfo{journal}{\emph{IEEE transactions on evolutionary
  computation}} \bibinfo{volume}{10}, \bibinfo{number}{4}
  (\bibinfo{year}{2006}), \bibinfo{pages}{459--472}.
\newblock


\bibitem[Blackwell et~al\mbox{.}(2008)]%
        {blackwell_particle_2008}
\bibfield{author}{\bibinfo{person}{Tim Blackwell}, \bibinfo{person}{J\"urgen
  Branke}, {and} \bibinfo{person}{Xiaodong Li}.}
  \bibinfo{year}{2008}\natexlab{}.
\newblock \showarticletitle{Particle {Swarms} for {Dynamic} {Optimization}
  {Problems}}.
\newblock In \bibinfo{booktitle}{\emph{Swarm {Intelligence}: {Introduction} and
  {Applications}}}, \bibfield{editor}{\bibinfo{person}{Christian Blum} {and}
  \bibinfo{person}{Daniel Merkle}} (Eds.). \bibinfo{publisher}{Springer},
  \bibinfo{address}{Berlin, Heidelberg}, \bibinfo{pages}{193--217}.
\newblock
\showISBNx{978-3-540-74089-6}
\href{https://doi.org/10.1007/978-3-540-74089-6_6}{doi:\nolinkurl{10.1007/978-3-540-74089-6_6}}


\bibitem[Boulesnane and Meshoul(2021)]%
        {boulesnane2021we}
\bibfield{author}{\bibinfo{person}{Abdennour Boulesnane} {and}
  \bibinfo{person}{Souham Meshoul}.} \bibinfo{year}{2021}\natexlab{}.
\newblock \showarticletitle{Do we need change detection for dynamic
  optimization problems?: A survey}. In \bibinfo{booktitle}{\emph{International
  Conference on Artificial Intelligence and its Applications}}. Springer,
  \bibinfo{pages}{132--142}.
\newblock


\bibitem[Branke(1999)]%
        {branke_memory_1999}
\bibfield{author}{\bibinfo{person}{J. Branke}.}
  \bibinfo{year}{1999}\natexlab{}.
\newblock \showarticletitle{Memory enhanced evolutionary algorithms for
  changing optimization problems}. In \bibinfo{booktitle}{\emph{Proceedings of
  the 1999 {Congress} on {Evolutionary} {Computation}-{CEC99} ({Cat}. {No}.
  {99TH8406})}}, Vol.~\bibinfo{volume}{3}. \bibinfo{pages}{1875--1882 Vol. 3}.
\newblock
\href{https://doi.org/10.1109/CEC.1999.785502}{doi:\nolinkurl{10.1109/CEC.1999.785502}}


\bibitem[Branke(2000)]%
        {branke_opt}
\bibfield{author}{\bibinfo{person}{J{\"{u}}rgen Branke}.}
  \bibinfo{year}{2000}\natexlab{}.
\newblock \emph{\bibinfo{title}{Evolutionary optimization in dynamic
  environments}}.
\newblock \bibinfo{thesistype}{Ph.\,D. Dissertation}.
  \bibinfo{school}{Universit{\"{a}}t Karlsruhe}.
\newblock


\bibitem[Branke et~al\mbox{.}(2000)]%
        {branke2000multi}
\bibfield{author}{\bibinfo{person}{J{\"u}rgen Branke}, \bibinfo{person}{Thomas
  Kau{\ss}ler}, \bibinfo{person}{Christian Smidt}, {and}
  \bibinfo{person}{Hartmut Schmeck}.} \bibinfo{year}{2000}\natexlab{}.
\newblock \showarticletitle{A multi-population approach to dynamic optimization
  problems}. In \bibinfo{booktitle}{\emph{Evolutionary Design and Manufacture:
  Selected Papers from ACDM’00}}. Springer, \bibinfo{pages}{299--307}.
\newblock


\bibitem[Branke and Schmeck(2003)]%
        {branke_designing_2003}
\bibfield{author}{\bibinfo{person}{J\"urgen Branke} {and}
  \bibinfo{person}{Hartmut Schmeck}.} \bibinfo{year}{2003}\natexlab{}.
\newblock \showarticletitle{Designing {Evolutionary} {Algorithms} for {Dynamic}
  {Optimization} {Problems}}.
\newblock In \bibinfo{booktitle}{\emph{Advances in {Evolutionary} {Computing}:
  {Theory} and {Applications}}}, \bibfield{editor}{\bibinfo{person}{Ashish
  Ghosh} {and} \bibinfo{person}{Shigeyoshi Tsutsui}} (Eds.).
  \bibinfo{publisher}{Springer}, \bibinfo{address}{Berlin, Heidelberg},
  \bibinfo{pages}{239--262}.
\newblock
\showISBNx{978-3-642-18965-4}
\href{https://doi.org/10.1007/978-3-642-18965-4_9}{doi:\nolinkurl{10.1007/978-3-642-18965-4_9}}


\bibitem[Brest et~al\mbox{.}(2009)]%
        {brest_2009_adaptive}
\bibfield{author}{\bibinfo{person}{Janez Brest}, \bibinfo{person}{Ales Zamuda},
  \bibinfo{person}{Borko Boskovic}, \bibinfo{person}{Mirjam~Sepesy Maucec},
  {and} \bibinfo{person}{Viljem Zumer}.} \bibinfo{year}{2009}\natexlab{}.
\newblock \showarticletitle{Dynamic optimization using Self-Adaptive
  Differential Evolution}. In \bibinfo{booktitle}{\emph{2009 IEEE Congress on
  Evolutionary Computation}}. \bibinfo{pages}{415--422}.
\newblock
\showISSN{1941-0026}
\href{https://doi.org/10.1109/CEC.2009.4982976}{doi:\nolinkurl{10.1109/CEC.2009.4982976}}


\bibitem[Chakraborty and Kar(2017)]%
        {chakraborty2017swarm}
\bibfield{author}{\bibinfo{person}{Amrita Chakraborty} {and}
  \bibinfo{person}{Arpan~Kumar Kar}.} \bibinfo{year}{2017}\natexlab{}.
\newblock \showarticletitle{Swarm intelligence: A review of algorithms}.
\newblock \bibinfo{journal}{\emph{Nature-inspired computing and optimization:
  Theory and applications}} (\bibinfo{year}{2017}), \bibinfo{pages}{475--494}.
\newblock


\bibitem[Coello(2007)]%
        {coello2007evolutionary}
\bibfield{author}{\bibinfo{person}{Carlos A~Coello Coello}.}
  \bibinfo{year}{2007}\natexlab{}.
\newblock \bibinfo{booktitle}{\emph{Evolutionary algorithms for solving
  multi-objective problems}}.
\newblock \bibinfo{publisher}{Springer}.
\newblock


\bibitem[{Danial Yazdani} et~al\mbox{.}(2020)]%
        {danial_yazdani_scaling_2020}
\bibfield{author}{\bibinfo{person}{{Danial Yazdani}}, \bibinfo{person}{Danial
  Yazdani}, \bibinfo{person}{{Mohammad Nabi Omidvar}},
  \bibinfo{person}{Mohammad~Nabi Omidvar}, \bibinfo{person}{{J\"urgen Branke}},
  \bibinfo{person}{J\"urgen Branke}, \bibinfo{person}{{Trung Thành Nguyen}},
  \bibinfo{person}{Trung~Thanh Nguyen}, \bibinfo{person}{{Xin Yao}}, {and}
  \bibinfo{person}{Xin Yao}.} \bibinfo{year}{2020}\natexlab{}.
\newblock \showarticletitle{Scaling {Up} {Dynamic} {Optimization} {Problems}:
  {A} {Divide}-and-{Conquer} {Approach}}.
\newblock \bibinfo{journal}{\emph{IEEE Transactions on Evolutionary
  Computation}} \bibinfo{volume}{24}, \bibinfo{number}{1} (\bibinfo{date}{Feb.}
  \bibinfo{year}{2020}), \bibinfo{pages}{1--15}.
\newblock
\href{https://doi.org/10.1109/tevc.2019.2902626}{doi:\nolinkurl{10.1109/tevc.2019.2902626}}
\newblock
\shownote{MAG ID: 2919775045}.


\bibitem[Das et~al\mbox{.}(2011)]%
        {das2011real}
\bibfield{author}{\bibinfo{person}{Swagatam Das}, \bibinfo{person}{Sayan
  Maity}, \bibinfo{person}{Bo-Yang Qu}, {and}
  \bibinfo{person}{Ponnuthurai~Nagaratnam Suganthan}.}
  \bibinfo{year}{2011}\natexlab{}.
\newblock \showarticletitle{Real-parameter evolutionary multimodal
  optimization—A survey of the state-of-the-art}.
\newblock \bibinfo{journal}{\emph{Swarm and Evolutionary Computation}}
  \bibinfo{volume}{1}, \bibinfo{number}{2} (\bibinfo{year}{2011}),
  \bibinfo{pages}{71--88}.
\newblock


\bibitem[Du~Plessis and Engelbrecht(2013)]%
        {du2013differential}
\bibfield{author}{\bibinfo{person}{Mathys~C Du~Plessis} {and}
  \bibinfo{person}{Andries~P Engelbrecht}.} \bibinfo{year}{2013}\natexlab{}.
\newblock \showarticletitle{Differential evolution for dynamic environments
  with unknown numbers of optima}.
\newblock \bibinfo{journal}{\emph{Journal of Global Optimization}}
  \bibinfo{volume}{55} (\bibinfo{year}{2013}), \bibinfo{pages}{73--99}.
\newblock


\bibitem[Fonseca and Fleming(1995)]%
        {fonseca1995overview}
\bibfield{author}{\bibinfo{person}{Carlos~M Fonseca} {and}
  \bibinfo{person}{Peter~J Fleming}.} \bibinfo{year}{1995}\natexlab{}.
\newblock \showarticletitle{An overview of evolutionary algorithms in
  multiobjective optimization}.
\newblock \bibinfo{journal}{\emph{Evolutionary computation}}
  \bibinfo{volume}{3}, \bibinfo{number}{1} (\bibinfo{year}{1995}),
  \bibinfo{pages}{1--16}.
\newblock


\bibitem[Halder et~al\mbox{.}(2011)]%
        {halder_self-adaptive_2011}
\bibfield{author}{\bibinfo{person}{Udit Halder}, \bibinfo{person}{Dipankar
  Maity}, \bibinfo{person}{Preetam Dasgupta}, {and} \bibinfo{person}{Swagatam
  Das}.} \bibinfo{year}{2011}\natexlab{}.
\newblock \showarticletitle{Self-adaptive cluster-based differential evolution
  with an external archive for dynamic optimization problems}. In
  \bibinfo{booktitle}{\emph{Swarm, Evolutionary, and Memetic Computing: Second
  International Conference, SEMCCO 2011, Visakhapatnam, Andhra Pradesh, India,
  December 19-21, 2011, Proceedings, Part I 2}}. Springer,
  \bibinfo{pages}{19--26}.
\newblock


\bibitem[Hashemi and Meybodi(2009)]%
        {hashemi_cellular_2009}
\bibfield{author}{\bibinfo{person}{Ali~B. Hashemi} {and}
  \bibinfo{person}{Mohammad~Reza Meybodi}.} \bibinfo{year}{2009}\natexlab{}.
\newblock \showarticletitle{Cellular {PSO}: {A} {PSO} for {Dynamic}
  {Environments}}. In \bibinfo{booktitle}{\emph{Advances in {Computation} and
  {Intelligence}, 4th {International} {Symposium}, {ISICA} 2009, {Huangshi},
  {China}, {Ocotober} 23-25, 2009, {Proceedings}}}
  \emph{(\bibinfo{series}{Lecture {Notes} in {Computer} {Science}},
  Vol.~\bibinfo{volume}{5821})}, \bibfield{editor}{\bibinfo{person}{Zhihua
  Cai}, \bibinfo{person}{Zhenhua Li}, \bibinfo{person}{Zhuo Kang}, {and}
  \bibinfo{person}{Yong Liu}} (Eds.). \bibinfo{publisher}{Springer},
  \bibinfo{pages}{422--433}.
\newblock
\href{https://doi.org/10.1007/978-3-642-04843-2_45}{doi:\nolinkurl{10.1007/978-3-642-04843-2_45}}


\bibitem[Hatzakis and Wallace(2006)]%
        {hatzakis2006dynamic}
\bibfield{author}{\bibinfo{person}{Iason Hatzakis} {and} \bibinfo{person}{David
  Wallace}.} \bibinfo{year}{2006}\natexlab{}.
\newblock \showarticletitle{Dynamic multi-objective optimization with
  evolutionary algorithms: a forward-looking approach}. In
  \bibinfo{booktitle}{\emph{Proceedings of the 8th annual conference on Genetic
  and evolutionary computation}}. \bibinfo{pages}{1201--1208}.
\newblock


\bibitem[Ho et~al\mbox{.}(2004)]%
        {ho2004intelligent}
\bibfield{author}{\bibinfo{person}{Shinn-Ying Ho}, \bibinfo{person}{Li-Sun
  Shu}, {and} \bibinfo{person}{Jian-Hung Chen}.}
  \bibinfo{year}{2004}\natexlab{}.
\newblock \showarticletitle{Intelligent evolutionary algorithms for large
  parameter optimization problems}.
\newblock \bibinfo{journal}{\emph{IEEE Transactions on evolutionary
  computation}} \bibinfo{volume}{8}, \bibinfo{number}{6}
  (\bibinfo{year}{2004}), \bibinfo{pages}{522--541}.
\newblock


\bibitem[Hu and Eberhart(2002)]%
        {hu_2002_detection}
\bibfield{author}{\bibinfo{person}{Xiaohui Hu} {and}
  \bibinfo{person}{Russell~C. Eberhart}.} \bibinfo{year}{2002}\natexlab{}.
\newblock \showarticletitle{Adaptive particle swarm optimization: detection and
  response to dynamic systems}. In \bibinfo{booktitle}{\emph{Proceedings of the
  2002 Congress on Evolutionary Computation, {CEC} 2002, Honolulu, HI, USA, May
  12-17, 2002}}. \bibinfo{publisher}{{IEEE}}, \bibinfo{pages}{1666--1670}.
\newblock
\href{https://doi.org/10.1109/CEC.2002.1004492}{doi:\nolinkurl{10.1109/CEC.2002.1004492}}


\bibitem[Huang et~al\mbox{.}(2020)]%
        {huang_survey_2020}
\bibfield{author}{\bibinfo{person}{Changwu Huang}, \bibinfo{person}{Yuanxiang
  Li}, {and} \bibinfo{person}{Xin Yao}.} \bibinfo{year}{2020}\natexlab{}.
\newblock \showarticletitle{A {Survey} of {Automatic} {Parameter} {Tuning}
  {Methods} for {Metaheuristics}}.
\newblock \bibinfo{journal}{\emph{IEEE Transactions on Evolutionary
  Computation}} \bibinfo{volume}{24}, \bibinfo{number}{2}
  (\bibinfo{date}{April} \bibinfo{year}{2020}), \bibinfo{pages}{201--216}.
\newblock
\showISSN{1941-0026}
\href{https://doi.org/10.1109/TEVC.2019.2921598}{doi:\nolinkurl{10.1109/TEVC.2019.2921598}}
\newblock
\shownote{Conference Name: IEEE Transactions on Evolutionary Computation}.


\bibitem[Ibrahim et~al\mbox{.}(2021)]%
        {ibrahim_2021_wind}
\bibfield{author}{\bibinfo{person}{Abdelhameed Ibrahim},
  \bibinfo{person}{Seyedali Mirjalili}, \bibinfo{person}{M. El-Said},
  \bibinfo{person}{Sherif S.~M. Ghoneim}, \bibinfo{person}{Mosleh~M.
  Al-Harthi}, \bibinfo{person}{Tarek~F. Ibrahim}, {and}
  \bibinfo{person}{El-Sayed~M. El-Kenawy}.} \bibinfo{year}{2021}\natexlab{}.
\newblock \showarticletitle{Wind Speed Ensemble Forecasting Based on Deep
  Learning Using Adaptive Dynamic Optimization Algorithm}.
\newblock \bibinfo{journal}{\emph{IEEE Access}}  \bibinfo{volume}{9}
  (\bibinfo{year}{2021}), \bibinfo{pages}{125787--125804}.
\newblock
\showISSN{2169-3536}
\href{https://doi.org/10.1109/ACCESS.2021.3111408}{doi:\nolinkurl{10.1109/ACCESS.2021.3111408}}


\bibitem[Jatmiko et~al\mbox{.}(2007)]%
        {jatmiko_2007}
\bibfield{author}{\bibinfo{person}{Wisnu Jatmiko}, \bibinfo{person}{Kosuke
  Sekiyama}, {and} \bibinfo{person}{Toshio Fukuda}.}
  \bibinfo{year}{2007}\natexlab{}.
\newblock \showarticletitle{A pso-based mobile robot for odor source
  localization in dynamic advection-diffusion with obstacles environment:
  theory, simulation and measurement}.
\newblock \bibinfo{journal}{\emph{{IEEE} Comput. Intell. Mag.}}
  \bibinfo{volume}{2}, \bibinfo{number}{2} (\bibinfo{year}{2007}),
  \bibinfo{pages}{37--51}.
\newblock
\href{https://doi.org/10.1109/MCI.2007.353419}{doi:\nolinkurl{10.1109/MCI.2007.353419}}


\bibitem[Kalita and Singh(2020)]%
        {kalita_2020_svm}
\bibfield{author}{\bibinfo{person}{Dhruba~Jyoti Kalita} {and}
  \bibinfo{person}{Shailendra Singh}.} \bibinfo{year}{2020}\natexlab{}.
\newblock \showarticletitle{{SVM} Hyper-parameters optimization using quantized
  multi-PSO in dynamic environment}.
\newblock \bibinfo{journal}{\emph{Soft Comput.}} \bibinfo{volume}{24},
  \bibinfo{number}{2} (\bibinfo{year}{2020}), \bibinfo{pages}{1225--1241}.
\newblock
\href{https://doi.org/10.1007/S00500-019-03957-W}{doi:\nolinkurl{10.1007/S00500-019-03957-W}}


\bibitem[Kamosi et~al\mbox{.}(2010)]%
        {kamosi_2010_hibernating}
\bibfield{author}{\bibinfo{person}{Masoud Kamosi}, \bibinfo{person}{Ali~B.
  Hashemi}, {and} \bibinfo{person}{Mohammad~Reza Meybodi}.}
  \bibinfo{year}{2010}\natexlab{}.
\newblock \showarticletitle{A hibernating multi-swarm optimization algorithm
  for dynamic environments}. In \bibinfo{booktitle}{\emph{Second World Congress
  on Nature {\&} Biologically Inspired Computing, NaBIC 2010, 15-17 December
  2010, Kitakyushu, Japan}}, \bibfield{editor}{\bibinfo{person}{Hideyuki
  Takagi}, \bibinfo{person}{Ajith Abraham}, \bibinfo{person}{Mario
  K{\"{o}}ppen}, \bibinfo{person}{Kaori Yoshida}, {and}
  \bibinfo{person}{Andr{\'{e}} C. P. L.~F. de~Carvalho}} (Eds.).
  \bibinfo{publisher}{{IEEE}}, \bibinfo{pages}{363--369}.
\newblock
\href{https://doi.org/10.1109/NABIC.2010.5716372}{doi:\nolinkurl{10.1109/NABIC.2010.5716372}}


\bibitem[Kennedy and Eberhart(1995)]%
        {kennedy_1995_PSO}
\bibfield{author}{\bibinfo{person}{J. Kennedy} {and} \bibinfo{person}{R.
  Eberhart}.} \bibinfo{year}{1995}\natexlab{}.
\newblock \showarticletitle{Particle swarm optimization}. In
  \bibinfo{booktitle}{\emph{Proceedings of ICNN'95 - International Conference
  on Neural Networks}}, Vol.~\bibinfo{volume}{4}. \bibinfo{pages}{1942--1948
  vol.4}.
\newblock
\href{https://doi.org/10.1109/ICNN.1995.488968}{doi:\nolinkurl{10.1109/ICNN.1995.488968}}


\bibitem[Lampinen et~al\mbox{.}(2005)]%
        {lampinen2005differential}
\bibfield{author}{\bibinfo{person}{Jouni~A Lampinen},
  \bibinfo{person}{Kenneth~V Price}, {and} \bibinfo{person}{Rainer~M Storn}.}
  \bibinfo{year}{2005}\natexlab{}.
\newblock \bibinfo{booktitle}{\emph{Differential evolution}}.
\newblock \bibinfo{publisher}{Springer}.
\newblock


\bibitem[Li et~al\mbox{.}(2016)]%
        {li_adaptive_2016}
\bibfield{author}{\bibinfo{person}{Changhe Li}, \bibinfo{person}{Trung~Thanh
  Nguyen}, \bibinfo{person}{Ming Yang}, \bibinfo{person}{Michalis
  Mavrovouniotis}, {and} \bibinfo{person}{Shengxiang Yang}.}
  \bibinfo{year}{2016}\natexlab{}.
\newblock \showarticletitle{An {Adaptive} {Multipopulation} {Framework} for
  {Locating} and {Tracking} {Multiple} {Optima}}.
\newblock \bibinfo{journal}{\emph{IEEE Transactions on Evolutionary
  Computation}} \bibinfo{volume}{20}, \bibinfo{number}{4} (\bibinfo{date}{Aug.}
  \bibinfo{year}{2016}), \bibinfo{pages}{590--605}.
\newblock
\showISSN{1941-0026}
\href{https://doi.org/10.1109/TEVC.2015.2504383}{doi:\nolinkurl{10.1109/TEVC.2015.2504383}}
\newblock
\shownote{Conference Name: IEEE Transactions on Evolutionary Computation}.


\bibitem[Li and Yang(2009)]%
        {li_clustering_2009}
\bibfield{author}{\bibinfo{person}{Changhe Li} {and}
  \bibinfo{person}{Shengxiang Yang}.} \bibinfo{year}{2009}\natexlab{}.
\newblock \showarticletitle{A clustering particle swarm optimizer for dynamic
  optimization}. In \bibinfo{booktitle}{\emph{2009 {IEEE} {Congress} on
  {Evolutionary} {Computation}}}. \bibinfo{pages}{439--446}.
\newblock
\href{https://doi.org/10.1109/CEC.2009.4982979}{doi:\nolinkurl{10.1109/CEC.2009.4982979}}
\newblock
\shownote{ISSN: 1941-0026}.


\bibitem[Li and Yang(2012)]%
        {li_general_2012}
\bibfield{author}{\bibinfo{person}{Changhe Li} {and}
  \bibinfo{person}{Shengxiang Yang}.} \bibinfo{year}{2012}\natexlab{}.
\newblock \showarticletitle{A {General} {Framework} of {Multipopulation}
  {Methods} {With} {Clustering} in {Undetectable} {Dynamic} {Environments}}.
\newblock \bibinfo{journal}{\emph{IEEE Transactions on Evolutionary
  Computation}} \bibinfo{volume}{16}, \bibinfo{number}{4} (\bibinfo{date}{Aug.}
  \bibinfo{year}{2012}), \bibinfo{pages}{556--577}.
\newblock
\showISSN{1089-778X, 1089-778X, 1941-0026}
\href{https://doi.org/10.1109/TEVC.2011.2169966}{doi:\nolinkurl{10.1109/TEVC.2011.2169966}}


\bibitem[Li et~al\mbox{.}(2008)]%
        {li2008benchmark}
\bibfield{author}{\bibinfo{person}{Changhe Li}, \bibinfo{person}{Shengxiang
  Yang}, \bibinfo{person}{Trung-Thanh Nguyen}, \bibinfo{person}{E~Ling Yu},
  \bibinfo{person}{Xin Yao}, \bibinfo{person}{Yaochu Jin}, \bibinfo{person}{HG
  Beyer}, {and} \bibinfo{person}{Ponnuthurai~Nagaratnam Suganthan}.}
  \bibinfo{year}{2008}\natexlab{}.
\newblock \bibinfo{booktitle}{\emph{Benchmark generator for CEC 2009
  competition on dynamic optimization}}.
\newblock \bibinfo{type}{{T}echnical {R}eport}.
\newblock


\bibitem[Li et~al\mbox{.}(2014)]%
        {li_adaptive_2014}
\bibfield{author}{\bibinfo{person}{Changhe Li}, \bibinfo{person}{Shengxiang
  Yang}, {and} \bibinfo{person}{Ming Yang}.} \bibinfo{year}{2014}\natexlab{}.
\newblock \showarticletitle{An {Adaptive} {Multi}-{Swarm} {Optimizer} for
  {Dynamic} {Optimization} {Problems}}.
\newblock \bibinfo{journal}{\emph{Evolutionary Computation}}
  \bibinfo{volume}{22}, \bibinfo{number}{4} (\bibinfo{date}{Dec.}
  \bibinfo{year}{2014}), \bibinfo{pages}{559--594}.
\newblock
\showISSN{1063-6560}
\href{https://doi.org/10.1162/EVCO_a_00117}{doi:\nolinkurl{10.1162/EVCO_a_00117}}


\bibitem[Li et~al\mbox{.}(2006)]%
        {li_particle_2006}
\bibfield{author}{\bibinfo{person}{Xiaodong Li}, \bibinfo{person}{J\"urgen
  Branke}, {and} \bibinfo{person}{Tim Blackwell}.}
  \bibinfo{year}{2006}\natexlab{}.
\newblock \showarticletitle{Particle swarm with speciation and adaptation in a
  dynamic environment}. In \bibinfo{booktitle}{\emph{Proceedings of the 8th
  annual conference on {Genetic} and evolutionary computation}}
  \emph{(\bibinfo{series}{{GECCO} '06})}. \bibinfo{publisher}{Association for
  Computing Machinery}, \bibinfo{address}{New York, NY, USA},
  \bibinfo{pages}{51--58}.
\newblock
\showISBNx{978-1-59593-186-3}
\href{https://doi.org/10.1145/1143997.1144005}{doi:\nolinkurl{10.1145/1143997.1144005}}


\bibitem[Liu et~al\mbox{.}(2011)]%
        {liu2011contamination}
\bibfield{author}{\bibinfo{person}{Li Liu}, \bibinfo{person}{S~Ranji
  Ranjithan}, {and} \bibinfo{person}{G Mahinthakumar}.}
  \bibinfo{year}{2011}\natexlab{}.
\newblock \showarticletitle{Contamination source identification in water
  distribution systems using an adaptive dynamic optimization procedure}.
\newblock \bibinfo{journal}{\emph{Journal of Water Resources Planning and
  Management}} \bibinfo{volume}{137}, \bibinfo{number}{2}
  (\bibinfo{year}{2011}), \bibinfo{pages}{183--192}.
\newblock


\bibitem[Liu et~al\mbox{.}(2006)]%
        {liu2006adaptive}
\bibfield{author}{\bibinfo{person}{Li Liu}, \bibinfo{person}{Emily~M Zechman},
  \bibinfo{person}{E~Downey Brill, Jr}, \bibinfo{person}{G Mahinthakumar},
  \bibinfo{person}{S Ranjithan}, {and} \bibinfo{person}{James Uber}.}
  \bibinfo{year}{2006}\natexlab{}.
\newblock \showarticletitle{Adaptive contamination source identification in
  water distribution systems using an evolutionary algorithm-based dynamic
  optimization procedure}. In \bibinfo{booktitle}{\emph{Water Distribution
  Systems Analysis Symposium 2006}}. \bibinfo{pages}{1--9}.
\newblock


\bibitem[Lung and Dumitrescu(2007)]%
        {lung_2007_tracking}
\bibfield{author}{\bibinfo{person}{Rodica~Ioana Lung} {and}
  \bibinfo{person}{Dumitru Dumitrescu}.} \bibinfo{year}{2007}\natexlab{}.
\newblock \showarticletitle{A collaborative model for tracking optima in
  dynamic environments}. In \bibinfo{booktitle}{\emph{Proceedings of the {IEEE}
  Congress on Evolutionary Computation, {CEC} 2007, 25-28 September 2007,
  Singapore}}. \bibinfo{publisher}{{IEEE}}, \bibinfo{pages}{564--567}.
\newblock
\href{https://doi.org/10.1109/CEC.2007.4424520}{doi:\nolinkurl{10.1109/CEC.2007.4424520}}


\bibitem[Mann and Whitney(1947)]%
        {mann_test_1947}
\bibfield{author}{\bibinfo{person}{H.~B. Mann} {and} \bibinfo{person}{D.~R.
  Whitney}.} \bibinfo{year}{1947}\natexlab{}.
\newblock \showarticletitle{On a {Test} of {Whether} one of {Two} {Random}
  {Variables} is {Stochastically} {Larger} than the {Other}}.
\newblock \bibinfo{journal}{\emph{The Annals of Mathematical Statistics}}
  \bibinfo{volume}{18}, \bibinfo{number}{1} (\bibinfo{year}{1947}),
  \bibinfo{pages}{50--60}.
\newblock
\showISSN{0003-4851}
\urldef\tempurl%
\url{https://www.jstor.org/stable/2236101}
\showURL{%
\tempurl}
\newblock
\shownote{Publisher: Institute of Mathematical Statistics}.


\bibitem[Mavrovouniotis et~al\mbox{.}(2017)]%
        {mavrovouniotis_survey_2017}
\bibfield{author}{\bibinfo{person}{Michalis Mavrovouniotis},
  \bibinfo{person}{Changhe Li}, {and} \bibinfo{person}{Shengxiang Yang}.}
  \bibinfo{year}{2017}\natexlab{}.
\newblock \showarticletitle{A survey of swarm intelligence for dynamic
  optimization: {Algorithms} and applications}.
\newblock \bibinfo{journal}{\emph{Swarm and Evolutionary Computation}}
  \bibinfo{volume}{33} (\bibinfo{date}{April} \bibinfo{year}{2017}),
  \bibinfo{pages}{1--17}.
\newblock
\showISSN{2210-6502}
\href{https://doi.org/10.1016/j.swevo.2016.12.005}{doi:\nolinkurl{10.1016/j.swevo.2016.12.005}}


\bibitem[Mendes and Mohais(2005)]%
        {dynDE}
\bibfield{author}{\bibinfo{person}{Rui Mendes} {and} \bibinfo{person}{Arvind~S.
  Mohais}.} \bibinfo{year}{2005}\natexlab{}.
\newblock \showarticletitle{DynDE: a differential evolution for dynamic
  optimization problems}. In \bibinfo{booktitle}{\emph{Proceedings of the
  {IEEE} Congress on Evolutionary Computation, {CEC} 2005, 2-4 September 2005,
  Edinburgh, {UK}}}. \bibinfo{publisher}{{IEEE}}, \bibinfo{pages}{2808--2815}.
\newblock
\href{https://doi.org/10.1109/CEC.2005.1555047}{doi:\nolinkurl{10.1109/CEC.2005.1555047}}


\bibitem[Michalewicz and Schoenauer(1996)]%
        {michalewicz1996evolutionary}
\bibfield{author}{\bibinfo{person}{Zbigniew Michalewicz} {and}
  \bibinfo{person}{Marc Schoenauer}.} \bibinfo{year}{1996}\natexlab{}.
\newblock \showarticletitle{Evolutionary algorithms for constrained parameter
  optimization problems}.
\newblock \bibinfo{journal}{\emph{Evolutionary computation}}
  \bibinfo{volume}{4}, \bibinfo{number}{1} (\bibinfo{year}{1996}),
  \bibinfo{pages}{1--32}.
\newblock


\bibitem[Mukherjee et~al\mbox{.}(2014)]%
        {mukherjee_2014_crowding}
\bibfield{author}{\bibinfo{person}{Rohan Mukherjee},
  \bibinfo{person}{Gyana~Ranjan Patra}, \bibinfo{person}{Rupam Kundu}, {and}
  \bibinfo{person}{Swagatam Das}.} \bibinfo{year}{2014}\natexlab{}.
\newblock \showarticletitle{Cluster-based differential evolution with Crowding
  Archive for niching in dynamic environments}.
\newblock \bibinfo{journal}{\emph{Information Sciences}}  \bibinfo{volume}{267}
  (\bibinfo{year}{2014}), \bibinfo{pages}{58--82}.
\newblock
\showISSN{0020-0255}
\href{https://doi.org/10.1016/j.ins.2013.11.025}{doi:\nolinkurl{10.1016/j.ins.2013.11.025}}


\bibitem[Nasiri and Meybodi(2016)]%
        {nasiri_2016_firefly}
\bibfield{author}{\bibinfo{person}{Babak Nasiri} {and}
  \bibinfo{person}{Mohammad~Reza Meybodi}.} \bibinfo{year}{2016}\natexlab{}.
\newblock \showarticletitle{History-driven firefly algorithm for optimisation
  in dynamic and uncertain environments}.
\newblock \bibinfo{journal}{\emph{Int. J. Bio Inspired Comput.}}
  \bibinfo{volume}{8}, \bibinfo{number}{5} (\bibinfo{year}{2016}),
  \bibinfo{pages}{326--339}.
\newblock
\href{https://doi.org/10.1504/IJBIC.2016.10000417}{doi:\nolinkurl{10.1504/IJBIC.2016.10000417}}


\bibitem[Nguyen(2011)]%
        {nguyen2011continuous}
\bibfield{author}{\bibinfo{person}{Trung~Thanh Nguyen}.}
  \bibinfo{year}{2011}\natexlab{}.
\newblock \emph{\bibinfo{title}{Continuous dynamic optimisation using
  evolutionary algorithms}}.
\newblock \bibinfo{thesistype}{Ph.\,D. Dissertation}.
  \bibinfo{school}{University of Birmingham}.
\newblock


\bibitem[Nguyen et~al\mbox{.}(2013)]%
        {nguyen2013solving}
\bibfield{author}{\bibinfo{person}{Trung~Thanh Nguyen}, \bibinfo{person}{Ian
  Jenkinson}, {and} \bibinfo{person}{Zaili Yang}.}
  \bibinfo{year}{2013}\natexlab{}.
\newblock \showarticletitle{Solving dynamic optimisation problems by combining
  evolutionary algorithms with KD-tree}. In \bibinfo{booktitle}{\emph{2013
  International Conference on Soft Computing and Pattern Recognition
  (SoCPaR)}}. IEEE, \bibinfo{pages}{247--252}.
\newblock


\bibitem[Nguyen et~al\mbox{.}(2012)]%
        {nguyen_evolutionary_2012}
\bibfield{author}{\bibinfo{person}{Trung~Thanh Nguyen},
  \bibinfo{person}{Shengxiang Yang}, {and} \bibinfo{person}{J\"urgen Branke}.}
  \bibinfo{year}{2012}\natexlab{}.
\newblock \showarticletitle{Evolutionary dynamic optimization: {A} survey of
  the state of the art}.
\newblock \bibinfo{journal}{\emph{Swarm Evol. Comput.}}  \bibinfo{volume}{6}
  (\bibinfo{year}{2012}), \bibinfo{pages}{1--24}.
\newblock
\href{https://doi.org/10.1016/J.SWEVO.2012.05.001}{doi:\nolinkurl{10.1016/J.SWEVO.2012.05.001}}


\bibitem[Nguyen and Yao(2012)]%
        {nguyen_2012_challenges}
\bibfield{author}{\bibinfo{person}{Trung~Thanh Nguyen} {and}
  \bibinfo{person}{Xin Yao}.} \bibinfo{year}{2012}\natexlab{}.
\newblock \showarticletitle{Continuous Dynamic Constrained Optimization—The
  Challenges}.
\newblock \bibinfo{journal}{\emph{IEEE Transactions on Evolutionary
  Computation}} \bibinfo{volume}{16}, \bibinfo{number}{6} (\bibinfo{date}{Dec}
  \bibinfo{year}{2012}), \bibinfo{pages}{769--786}.
\newblock
\showISSN{1941-0026}
\href{https://doi.org/10.1109/TEVC.2011.2180533}{doi:\nolinkurl{10.1109/TEVC.2011.2180533}}


\bibitem[Novoa{-}Hern{\'{a}}ndez et~al\mbox{.}(2010)]%
        {novoa_2010_improvement}
\bibfield{author}{\bibinfo{person}{Pavel Novoa{-}Hern{\'{a}}ndez},
  \bibinfo{person}{David~A. Pelta}, {and} \bibinfo{person}{Carlos~Cruz
  Corona}.} \bibinfo{year}{2010}\natexlab{}.
\newblock \showarticletitle{Improvement Strategies for Multi-swarm {PSO} in
  Dynamic Environments}. In \bibinfo{booktitle}{\emph{Nature Inspired
  Cooperative Strategies for Optimization, {NICSO} 2010, May 12-14, 2010,
  Granada, Spain}} \emph{(\bibinfo{series}{Studies in Computational
  Intelligence}, Vol.~\bibinfo{volume}{284})}. \bibinfo{publisher}{Springer},
  \bibinfo{pages}{371--383}.
\newblock
\href{https://doi.org/10.1007/978-3-642-12538-6\_31}{doi:\nolinkurl{10.1007/978-3-642-12538-6\_31}}


\bibitem[Parrott and Li(2004)]%
        {parrott2004particle}
\bibfield{author}{\bibinfo{person}{Daniel Parrott} {and}
  \bibinfo{person}{Xiaodong Li}.} \bibinfo{year}{2004}\natexlab{}.
\newblock \showarticletitle{A particle swarm model for tracking multiple peaks
  in a dynamic environment using speciation}. In
  \bibinfo{booktitle}{\emph{Proceedings of the 2004 congress on evolutionary
  computation (IEEE Cat. No. 04TH8753)}}, Vol.~\bibinfo{volume}{1}. IEEE,
  \bibinfo{pages}{98--103}.
\newblock


\bibitem[Parrott and Li(2006)]%
        {parrott_locating_2006}
\bibfield{author}{\bibinfo{person}{D. Parrott} {and} \bibinfo{person}{Xiaodong
  Li}.} \bibinfo{year}{2006}\natexlab{}.
\newblock \showarticletitle{Locating and tracking multiple dynamic optima by a
  particle swarm model using speciation}.
\newblock \bibinfo{journal}{\emph{IEEE Transactions on Evolutionary
  Computation}} \bibinfo{volume}{10}, \bibinfo{number}{4} (\bibinfo{date}{Aug.}
  \bibinfo{year}{2006}), \bibinfo{pages}{440--458}.
\newblock
\showISSN{1941-0026}
\href{https://doi.org/10.1109/TEVC.2005.859468}{doi:\nolinkurl{10.1109/TEVC.2005.859468}}
\newblock
\shownote{Conference Name: IEEE Transactions on Evolutionary Computation}.


\bibitem[Peng et~al\mbox{.}(2024)]%
        {peng_edolab_2024}
\bibfield{author}{\bibinfo{person}{Mai Peng}, \bibinfo{person}{Delaram
  Yazdani}, \bibinfo{person}{Zeneng She}, \bibinfo{person}{Danial Yazdani},
  \bibinfo{person}{Wenjian Luo}, \bibinfo{person}{Changhe Li},
  \bibinfo{person}{Juergen Branke}, \bibinfo{person}{Trung~Thanh Nguyen},
  \bibinfo{person}{Amir~H. Gandomi}, \bibinfo{person}{Shengxiang Yang},
  \bibinfo{person}{Yaochu Jin}, {and} \bibinfo{person}{Xin Yao}.}
  \bibinfo{year}{2024}\natexlab{}.
\newblock \bibinfo{title}{{EDOLAB}: {An} {Open}-{Source} {Platform} for
  {Education} and {Experimentation} with {Evolutionary} {Dynamic}
  {Optimization} {Algorithms}}.
\newblock
\href{https://doi.org/10.48550/arXiv.2308.12644}{doi:\nolinkurl{10.48550/arXiv.2308.12644}}
\newblock
\shownote{arXiv:2308.12644 [cs]}.


\bibitem[Richter(2009)]%
        {richter_2009_detecting}
\bibfield{author}{\bibinfo{person}{Hendrik Richter}.}
  \bibinfo{year}{2009}\natexlab{}.
\newblock \showarticletitle{Detecting change in dynamic fitness landscapes}. In
  \bibinfo{booktitle}{\emph{2009 IEEE Congress on Evolutionary Computation}}.
  \bibinfo{pages}{1613--1620}.
\newblock
\showISSN{1941-0026}
\href{https://doi.org/10.1109/CEC.2009.4983135}{doi:\nolinkurl{10.1109/CEC.2009.4983135}}


\bibitem[Trojanowski(2009)]%
        {trojanowski2009properties}
\bibfield{author}{\bibinfo{person}{Krzysztof Trojanowski}.}
  \bibinfo{year}{2009}\natexlab{}.
\newblock \showarticletitle{Properties of quantum particles in multi-swarms for
  dynamic optimization}.
\newblock \bibinfo{journal}{\emph{Fundamenta Informaticae}}
  \bibinfo{volume}{95}, \bibinfo{number}{2-3} (\bibinfo{year}{2009}),
  \bibinfo{pages}{349--380}.
\newblock


\bibitem[Trojanowski and Michalewicz(1999)]%
        {trojanowski1999searching}
\bibfield{author}{\bibinfo{person}{Krzysztof Trojanowski} {and}
  \bibinfo{person}{Zbigniew Michalewicz}.} \bibinfo{year}{1999}\natexlab{}.
\newblock \showarticletitle{Searching for optima in non-stationary
  environments}. In \bibinfo{booktitle}{\emph{Proceedings of the 1999 congress
  on evolutionary computation-CEC99 (Cat. No. 99TH8406)}},
  Vol.~\bibinfo{volume}{3}. IEEE, \bibinfo{pages}{1843--1850}.
\newblock


\bibitem[Wang et~al\mbox{.}(2007)]%
        {wang2007triggered}
\bibfield{author}{\bibinfo{person}{Hongfeng Wang}, \bibinfo{person}{Dingwei
  Wang}, {and} \bibinfo{person}{Shengxiang Yang}.}
  \bibinfo{year}{2007}\natexlab{}.
\newblock \showarticletitle{Triggered memory-based swarm optimization in
  dynamic environments}. In \bibinfo{booktitle}{\emph{Applications of
  Evolutionary Computing: EvoWorkshops 2007: EvoCoMnet, EvoFIN, EvoIASP,
  EvoINTERACTION, EvoMUSART, EvoSTOC and EvoTransLog. Proceedings}}. Springer,
  \bibinfo{pages}{637--646}.
\newblock


\bibitem[Wang et~al\mbox{.}(2011)]%
        {wang_2011_dispatch}
\bibfield{author}{\bibinfo{person}{Ying Wang}, \bibinfo{person}{Jianzhong
  Zhou}, \bibinfo{person}{Youlin Lu}, \bibinfo{person}{Hui Qin}, {and}
  \bibinfo{person}{Yongqiang Wang}.} \bibinfo{year}{2011}\natexlab{}.
\newblock \showarticletitle{Chaotic self-adaptive particle swarm optimization
  algorithm for dynamic economic dispatch problem with valve-point effects}.
\newblock \bibinfo{journal}{\emph{Expert Syst. Appl.}} \bibinfo{volume}{38},
  \bibinfo{number}{11} (\bibinfo{year}{2011}), \bibinfo{pages}{14231--14237}.
\newblock
\href{https://doi.org/10.1016/J.ESWA.2011.04.236}{doi:\nolinkurl{10.1016/J.ESWA.2011.04.236}}


\bibitem[Weise(2009)]%
        {weise2009global}
\bibfield{author}{\bibinfo{person}{Thomas Weise}.}
  \bibinfo{year}{2009}\natexlab{}.
\newblock \showarticletitle{Global optimization algorithms-theory and
  application}.
\newblock \bibinfo{journal}{\emph{Self-Published Thomas Weise}}
  \bibinfo{volume}{361} (\bibinfo{year}{2009}), \bibinfo{pages}{153}.
\newblock


\bibitem[Yang and Li(2010)]%
        {yang_2010_tracking}
\bibfield{author}{\bibinfo{person}{Shengxiang Yang} {and}
  \bibinfo{person}{Changhe Li}.} \bibinfo{year}{2010}\natexlab{}.
\newblock \showarticletitle{A Clustering Particle Swarm Optimizer for Locating
  and Tracking Multiple Optima in Dynamic Environments}.
\newblock \bibinfo{journal}{\emph{{IEEE} Trans. Evol. Comput.}}
  \bibinfo{volume}{14}, \bibinfo{number}{6} (\bibinfo{year}{2010}),
  \bibinfo{pages}{959--974}.
\newblock
\href{https://doi.org/10.1109/TEVC.2010.2046667}{doi:\nolinkurl{10.1109/TEVC.2010.2046667}}


\bibitem[Yazdani et~al\mbox{.}(2021a)]%
        {benchmark_2024}
\bibfield{author}{\bibinfo{person}{Danial Yazdani},
  \bibinfo{person}{J{\"{u}}rgen Branke}, \bibinfo{person}{Mohammad~Nabi
  Omidvar}, \bibinfo{person}{Changhe Li}, \bibinfo{person}{Michalis
  Mavrovouniotis}, \bibinfo{person}{Trung~Thanh Nguyen},
  \bibinfo{person}{Shengxiang Yang}, {and} \bibinfo{person}{Xin Yao}.}
  \bibinfo{year}{2021}\natexlab{a}.
\newblock \showarticletitle{Generalized Moving Peaks Benchmark}.
\newblock \bibinfo{journal}{\emph{CoRR}}  \bibinfo{volume}{abs/2106.06174}
  (\bibinfo{year}{2021}).
\newblock
\showeprint[arXiv]{2106.06174}
\urldef\tempurl%
\url{https://arxiv.org/abs/2106.06174}
\showURL{%
\tempurl}


\bibitem[Yazdani et~al\mbox{.}(2022)]%
        {yazdani_adaptive_2022}
\bibfield{author}{\bibinfo{person}{Danial Yazdani}, \bibinfo{person}{Ran
  Cheng}, \bibinfo{person}{Cheng He}, {and} \bibinfo{person}{J\"urgen Branke}.}
  \bibinfo{year}{2022}\natexlab{}.
\newblock \showarticletitle{Adaptive {Control} of {Subpopulations} in
  {Evolutionary} {Dynamic} {Optimization}}.
\newblock \bibinfo{journal}{\emph{IEEE Transactions on Cybernetics}}
  \bibinfo{volume}{52}, \bibinfo{number}{7} (\bibinfo{date}{July}
  \bibinfo{year}{2022}), \bibinfo{pages}{6476--6489}.
\newblock
\showISSN{2168-2275}
\href{https://doi.org/10.1109/TCYB.2020.3036100}{doi:\nolinkurl{10.1109/TCYB.2020.3036100}}
\newblock
\shownote{Conference Name: IEEE Transactions on Cybernetics}.


\bibitem[Yazdani et~al\mbox{.}(2021b)]%
        {yazdani_2021_A}
\bibfield{author}{\bibinfo{person}{Danial Yazdani}, \bibinfo{person}{Ran
  Cheng}, \bibinfo{person}{Donya Yazdani}, \bibinfo{person}{J\"urgen Branke},
  \bibinfo{person}{Yaochu Jin}, {and} \bibinfo{person}{Xin Yao}.}
  \bibinfo{year}{2021}\natexlab{b}.
\newblock \showarticletitle{A Survey of Evolutionary Continuous Dynamic
  Optimization Over Two Decades - Part {A}}.
\newblock \bibinfo{journal}{\emph{{IEEE} Trans. Evol. Comput.}}
  \bibinfo{volume}{25}, \bibinfo{number}{4} (\bibinfo{year}{2021}),
  \bibinfo{pages}{609--629}.
\newblock
\href{https://doi.org/10.1109/TEVC.2021.3060014}{doi:\nolinkurl{10.1109/TEVC.2021.3060014}}


\bibitem[Yazdani et~al\mbox{.}(2021c)]%
        {yazdani_survey_2021}
\bibfield{author}{\bibinfo{person}{Danial Yazdani}, \bibinfo{person}{Ran
  Cheng}, \bibinfo{person}{Donya Yazdani}, \bibinfo{person}{J\"urgen Branke},
  \bibinfo{person}{Yaochu Jin}, {and} \bibinfo{person}{Xin Yao}.}
  \bibinfo{year}{2021}\natexlab{c}.
\newblock \showarticletitle{A {Survey} of {Evolutionary} {Continuous} {Dynamic}
  {Optimization} {Over} {Two} {Decades} - {Part} {B}}.
\newblock \bibinfo{journal}{\emph{IEEE Trans. Evol. Comput.}}
  \bibinfo{volume}{25}, \bibinfo{number}{4} (\bibinfo{year}{2021}),
  \bibinfo{pages}{630--650}.
\newblock
\href{https://doi.org/10.1109/TEVC.2021.3060012}{doi:\nolinkurl{10.1109/TEVC.2021.3060012}}


\bibitem[Yazdani et~al\mbox{.}(2024)]%
        {yazdani2024competitiondynamicoptimizationproblems}
\bibfield{author}{\bibinfo{person}{Danial Yazdani}, \bibinfo{person}{Michalis
  Mavrovouniotis}, \bibinfo{person}{Changhe Li}, \bibinfo{person}{Guoyu Chen},
  \bibinfo{person}{Wenjian Luo}, \bibinfo{person}{Mohammad~Nabi Omidvar},
  \bibinfo{person}{Juergen Branke}, \bibinfo{person}{Shengxiang Yang}, {and}
  \bibinfo{person}{Xin Yao}.} \bibinfo{year}{2024}\natexlab{}.
\newblock \bibinfo{title}{Competition on Dynamic Optimization Problems
  Generated by Generalized Moving Peaks Benchmark (GMPB)}.
\newblock
\showeprint[arxiv]{2106.06174}~[cs.NE]
\urldef\tempurl%
\url{https://arxiv.org/abs/2106.06174}
\showURL{%
\tempurl}


\bibitem[Yazdani et~al\mbox{.}(2013a)]%
        {yazdani2013novel}
\bibfield{author}{\bibinfo{person}{Danial Yazdani}, \bibinfo{person}{Babak
  Nasiri}, \bibinfo{person}{Alireza Sepas-Moghaddam}, {and}
  \bibinfo{person}{Mohammad~Reza Meybodi}.} \bibinfo{year}{2013}\natexlab{a}.
\newblock \showarticletitle{A novel multi-swarm algorithm for optimization in
  dynamic environments based on particle swarm optimization}.
\newblock \bibinfo{journal}{\emph{Applied Soft Computing}}
  \bibinfo{volume}{13}, \bibinfo{number}{4} (\bibinfo{year}{2013}),
  \bibinfo{pages}{2144--2158}.
\newblock


\bibitem[Yazdani et~al\mbox{.}(2013b)]%
        {yazdani_2013_novel}
\bibfield{author}{\bibinfo{person}{Danial Yazdani}, \bibinfo{person}{Babak
  Nasiri}, \bibinfo{person}{Alireza Sepas{-}Moghaddam}, {and}
  \bibinfo{person}{Mohammad~Reza Meybodi}.} \bibinfo{year}{2013}\natexlab{b}.
\newblock \showarticletitle{A novel multi-swarm algorithm for optimization in
  dynamic environments based on particle swarm optimization}.
\newblock \bibinfo{journal}{\emph{Appl. Soft Comput.}} \bibinfo{volume}{13},
  \bibinfo{number}{4} (\bibinfo{year}{2013}), \bibinfo{pages}{2144--2158}.
\newblock
\href{https://doi.org/10.1016/J.ASOC.2012.12.020}{doi:\nolinkurl{10.1016/J.ASOC.2012.12.020}}


\bibitem[Yazdani et~al\mbox{.}(2020)]%
        {yazdani2020benchmarking}
\bibfield{author}{\bibinfo{person}{Danial Yazdani},
  \bibinfo{person}{Mohammad~Nabi Omidvar}, \bibinfo{person}{Ran Cheng},
  \bibinfo{person}{J{\"u}rgen Branke}, \bibinfo{person}{Trung~Thanh Nguyen},
  {and} \bibinfo{person}{Xin Yao}.} \bibinfo{year}{2020}\natexlab{}.
\newblock \showarticletitle{Benchmarking continuous dynamic optimization:
  Survey and generalized test suite}.
\newblock \bibinfo{journal}{\emph{IEEE transactions on cybernetics}}
  \bibinfo{volume}{52}, \bibinfo{number}{5} (\bibinfo{year}{2020}),
  \bibinfo{pages}{3380--3393}.
\newblock


\bibitem[Yazdani et~al\mbox{.}(2023)]%
        {yazdani_species-based_2023}
\bibfield{author}{\bibinfo{person}{Delaram Yazdani}, \bibinfo{person}{Danial
  Yazdani}, \bibinfo{person}{Donya Yazdani}, \bibinfo{person}{Mohammad~Nabi
  Omidvar}, \bibinfo{person}{Amir~H. Gandomi}, {and} \bibinfo{person}{Xin
  Yao}.} \bibinfo{year}{2023}\natexlab{}.
\newblock \showarticletitle{A {Species}-based {Particle} {Swarm} {Optimization}
  with {Adaptive} {Population} {Size} and {Deactivation} of {Species} for
  {Dynamic} {Optimization} {Problems}}.
\newblock \bibinfo{journal}{\emph{ACM Transactions on Evolutionary Learning and
  Optimization}} \bibinfo{volume}{3}, \bibinfo{number}{4}
  (\bibinfo{year}{2023}), \bibinfo{pages}{14:1--14:25}.
\newblock
\href{https://doi.org/10.1145/3604812}{doi:\nolinkurl{10.1145/3604812}}


\end{thebibliography}



\end{document}